\documentclass[journal]{IEEEtran}

\usepackage{amsmath}
\usepackage{graphicx}
\usepackage{verbatim}
\usepackage{siunitx}
\usepackage{multirow}
\usepackage{colortbl}
\usepackage{xcolor}
\usepackage{amsfonts}
\usepackage{mathtools}
\usepackage{xr}
\usepackage{cite}

\ifCLASSINFOpdf
\else
\fi

\begin{document}

\title{Representative period selection for power system planning using autoencoder-based dimensionality reduction}

\author{\IEEEauthorblockN{Marc Barbar\IEEEauthorrefmark{1},
Dharik S. Mallapragada\IEEEauthorrefmark{1}}\\
\IEEEauthorblockA{\IEEEauthorrefmark{1}MIT Energy Initiative, Massachusetts Institute of Technology, Cambridge, MA 02139 USA}
\thanks{Corresponding author: D.S. Mallapragada (email: dharik@mit.edu).}}

\maketitle

\begin{abstract}
Power sector capacity expansion models (CEMs) that are used for studying future low-carbon grid scenarios must incorporate detailed representation of grid operations.
Often CEMs are formulated to model grid operations over representative periods that are sampled from the original input data using clustering algorithms. However, such representative period selection (RPS) methods are limited by the declining efficacy of the clustering algorithm with increasing dimensionality of the input data and do not consider the relative importance of input data variations on CEM outcomes. Here, we propose a RPS method that addresses these limitations by incorporating dimensionality reduction, accomplished via neural network based autoencoders, prior to clustering. Such dimensionality reduction not only improves the performance of the clustering algorithm, but also facilitates using additional features, such as estimated outputs produced from parallel solutions of simplified versions of the CEM for each disjoint period in the input data (e.g. 1 week). The impact of incorporating dimensionality reduction as part of RPS methods is quantified through the error in outcomes of the corresponding reduced-space CEM vs. the full space CEM. Extensive numerical experimentation across various networks and range of technology and policy scenarios establish the
superiority of the dimensionality-reduction based RPS methods.

\end{abstract}


%
\IEEEpeerreviewmaketitle





\section{Introduction}
\IEEEPARstart{L}{east-cost} planning of deeply decarbonized bulk power systems requires contending with the unique operational attributes of the demand and supply-side resources that play a dominant role in such systems. These include: a) variable renewable energy sources (VRE) whose output varies across multiple time-scales; b) new flexible loads such as heat pumps, electric vehicles and distributed energy resources that collectively increase demand variability and c) storage resources that store energy for various time scales, ranging from intra-day to seasons, and thus couple grid operations across these periods \cite{Sepulveda2021, Brown2020, Guerra2020}. Recognizing this need, a growing body of literature has focused on formulation and solution strategies for power system capacity expansion models (CEM) with improved spatial and temporal representation of grid operations \cite{MALLAPRAGADA2020115390, kuepper2020wind, teichgraeber2020extreme, tejada2018enhanced, 8017598, 9712383, 8536427, 8633397, poncelet2016selecting}. 

A typical CEM takes the viewpoint of central planner and minimizes the sum of investment cost and operational costs of electricity supply, network, storage and demand-side resources. The constraints of a CEM usually include operational constraints of each resource as well as the system including, supply-demand balance, reliability requirements, and environmental considerations such as carbon emissions limits \cite{KOLTSAKLIS2018563, PONCELET2016631}. The number of variables in the CEM scale linearly with the number of operational periods modeled; let $n$ be the total number of generation units, $l$ number of load zones and $t$ number of operational periods. Then the full-scale CEM is a problem of dimension $(n+l) \times t$. To maintain computational tractability, many CEMs evaluate grid operations over a few representative periods (e.g. days, weeks), often selected using one of the many known variations of time series clustering techniques (which we cover later in this section) \cite{wogrin2014new, MALLAPRAGADA2020115390, baumgartner2019rises3}. In the context of an electricity system, time series clustering usually face common challenges such as processing multidimensional and multivariate input data and handling hidden features that are only detected in the results \cite{TEICHGRAEBER2022111984}. Therefore, along with advancements in decomposition algorithms that allow for increasing temporal resolution of CEMs while maintaining computational tractability \cite{Lara2018, 8316946}, improvements in representative period selection (RPS) methods to exploit unique attributes of power sector-related time series data are useful. 

The problem of time series clustering for RPS finds applications in many domains outside power systems, and generally consists of four major components: 1) clustering algorithm, 2) prototype definition, 3) distance measurement, and 4) dimensionality reduction.

\begin{enumerate}

    \item \textbf{Clustering Algorithms:} Commonly-used time series clustering algorithms can be classified into six broad classes depending on the underlying logic used for grouping the elements (e.g. hour, day, week) of the time series into clusters: hierarchical (bottom-up and top-down) \cite{Macqueen67somemethods}, partition-based (e.g. \textit{k}-means, k-mediod) \cite{scikit-learn}, density-based (DBSCAN) \cite{10.5555/3001460.3001507}, grid-based (e.g., finite cell division) \cite{Sheikholeslami1998WaveClusterAM}, model-based (e.g., self-organizing maps) \cite{Fu2016PatternDF}, and multi-step (hybrid methods) \cite{Aghabozorgi2014}.
    
    \item \textbf{Prototype Definition:} The prototype of a cluster refers to the representation of the original time series in the reduced space. There are three main prototyping methods: 1) means and medoid \cite{Kaufman1990}, 2) temporal averaging (e.g., Dynamic Time Warping (DTW)) \cite{10.5555/3000292.3000335}, and 3) local search \cite{4761105}.
    
    \item \textbf{Distance Metric:} Time series clustering highly depends on the choice of the distance measure used for assigning the time series elements to clusters \cite{10.1145/331499.331504}. Commonly used distance metrics include 1) time-based metrics such as Euclidean, correlation, Fourier transform \cite{Keogh2003, Bagnall2005}, 2) shape-based metrics \cite{doi:10.1137/1.9781611972726.12}, and 3) change-based metrics (e.g., Hidden Markov Models) \cite{Wang2006}.
    
    \item \textbf{Dimensionality Reduction:} Unlike the other three components, which are essential for the clustering process, dimensionality reduction is an optional component. Broadly speaking, a clustering algorithm's performance is measured by how well it can assign a cluster label to a given dataset, commonly measured by the prototype definition and distance metric. Since a clustering algorithm's performance deteriorates when the size of the data increases \cite{scikit-learn}, dimensionality reduction downsizes the dataset used in the clustering process for better performance and lower computational burden. This improved performance comes at the expense of information loss in the dimensionality reduction process. Dimensionality reduction methods include 1) data-adaptive approaches \cite{Ratanamahatana2005}, 2) deterministic segmentation \cite{Lin2003}, 3) model-based (e.g., Auto-Regressive Moving Average) and \cite{Bagnall2006} 4) clipping \cite{Shieh2008iSAXIA}.

\end{enumerate}

The above-mentioned components of RPS have been heavily discussed in the literature in the context of electricity resource planning \cite{TEICHGRAEBER20191283, kuepper2020wind, teichgraeber2020extreme, TSO2020115190, 7527691, 8442580, 8369128}. Several studies have explored the problem of RPS for CEMs applied to power and/or energy systems \cite{hoffmann2020review, 7527691, 8442580}. Here, we summarize a few key papers that exemplify the general approaches. Authors in\cite{8334256, tejada2018enhanced} explore modeling representative days for optimal deployment of energy storage only. Probabilistic generation snapshots method to cluster representative weeks based on mean cluster values is discussed in \cite{Fitiwi2015}. Operational pattern similarity in temporal data to aggregate generation dispatch models is also proposed in some studies\cite{lythcke2016method, sun2019data} but without considering the temporal variations in supply-demand balance. Several papers have resorted to using variants of \textit{k}-means clustering on the original input data to select representative periods to be used in CEM evaluation \cite{8633397, MALLAPRAGADA2020115390, baumgartner2019rises3, LI2022107697}. Other papers have proposed RPS methods based on hierarchical clustering \cite{8017598, TSO2020115190} or simulation-based methods \cite{8610317, 8369128}. Other studies \cite{7527691, 8442580, TEICHGRAEBER2022111984} explore optimization-based approaches at a high computational cost that limits its scalability. On the other hand, data-driven approaches \cite{Tavakoli2020, Guo2017} have been proven to be effective in time series aggregation. However, the proposed methods are restricted to labeled data sets that are generally not applicable to power system planning because the time series data is not labeled and require \textit{a priori} knowledge of the distribution function for probabilistic assignment to a cluster based on the distance metric. Additionally, the performance of such methods is often measured by the error in reproducing the input data to a CEM, whereas in power systems it is the CEM output that determines the quality of the RPS. Recent work on learning-based methods \cite{8663347} has been proposed to evaluate RPS of wind resources but without any application to CEM of power systems.

In summary, the literature on RPS for power system CEMs has generally focused on using the original multi-variate input data directly in the clustering process without considering how well the input data approximates the original output (i.e. full-space problem). The contributions of this paper is twofold: First, we introduce a data pre-processing step involving nonlinear dimensionality reduction via neural networks. We call this process an \textit{autoencoder} and explain in detail in section \ref{method}. Second, we propose a technique to incorporate CEM output information in the clustering process. The resulting approach allows us to find a balance between loss of information during the dimensionlaity reduction process and the information loss incurred during the clustering process. Our approach identifies a set of representative periods and their corresponding input data that is used to formulate a reduced-space CEM (RCEM, i.e. the CEM with representative periods) whose outcomes are shown to approximate full-space CEM (FCEM, i.e. CEM using full-dimension input data) outcomes with an acceptable accuracy. The autoencoders identify both short (intra-day variations) and long-term features (seasonal variations) in the original multi-variate time series data and reduce the dimension of the data in a way that reduces the clustering loss. We perform numerical experiments to compare the performance of the proposed RPS method with two alternative methods from the literature which rely on k-means clustering but do not use dimensionality reduction. In all four approaches, we fix the prototype definition of the clustering process to be \textit{mediod}, use \textit{k}-means as the clustering algorithm and use Euclidean distance (L2 norm) as the distance metric. Overall, the main contributions of the paper to the field of power system planning models are as follows:
\begin{enumerate}
    \item use of autoencoders for dimensionality reduction as a pre-processing step in the RPS method.
    \item defining and incorporating features related to CEM outputs to be used in the clustering process.
    \item extensive numerical experimentation spanning different number of representative periods selected (4, 8, 20) for RCEM, different network sizes (1, 3, 8 bus) and cost assumptions (90 scenario runs for each network and representative period pair) for the case study.
\end{enumerate}

The remainder of the paper is structured as follows: section \ref{method} describes the methods, including the key assumptions and structure of the proposed methods and formulation of the CEM; section \ref{casestudy} details the case study input data we use for the comparative statistical analysis of the four alternative RPS methods; section \ref{results} describes results of the various RPS methods from the case study. Finally, section \ref{conclusion} concludes the paper, discusses the limitations and, provides future directions.

\section{Methods}\label{method}

\subsection{CEM model, input and output data}

To evaluate the proposed methods, we use a standard linear programming CEM formulation \cite{taylor2015convex, Jenkins2017, Knueven2020} that is fully described in Appendix \ref{si} section \ref{optmodel}. The model objective function (Eq. \ref{obj}) minimizes the total system cost which includes annualized generation expansion and, operational costs. The full-space CEM (FCEM) has $w_t = 1$ for every period $t$. On the other hand, the reduced-space CEM (RCEM) has time weight $w_t \geq 1$ depending on the RPS. The key operational constraints are: 1) hourly power balance constraint (Eq. \ref{dem}), 2) time-dependent capacity constraints for generation resources (Eq. \ref{vre}), 3) battery storage state of charging, energy and power capacity limits (Eq. \ref{stor1} - \ref{stor4}), 4) generation unit commitment (Eq. \ref{therm1} - \ref{therm1_2}), 5) generation minimum and maximum power output (Eq. \ref{therm2}, \ref{therm3}), 6) generation ramping limits (Eq. \ref{therm4}, \ref{therm5}), 7) direct current power flow approximation using line susceptance and bus angles (Eq. \ref{net1}), 8) network flow limits (Eq. \ref{net2}, \ref{net3}) and, 9) non-negativity constraints (Eq. \ref{nnc}). This formulation is used for both the FCEM and RCEM, with the following two caveats: First, the RCEM does not include inter-temporal constraints linking resource operations across representative periods. For the inter-temporal constraints related to ramping limits and storage inventory, we look back to the last sub-period of the same representative period to define the constraint for the first sub-period (see Eq. \ref{stor1_1}) \cite{MALLAPRAGADA2020115390, 8442580, poncelet2016selecting}. This modeling approximation implies that storage inventory across two representative periods is effectively decoupled, which is generally reasonable when considering short-duration storage technologies. Second, the RCEM computes annual operating costs based on scaling up operating costs of representative periods using weights ($w_t$) obtained from the clustering process that correspond to number of periods approximated by each representative period.

The solution of the CEM formulation yields a set of generation, energy and charge capacities ($\Omega^{size}_{a,z}$, $\Omega^{energy}_{a,z}$ and, $\Omega^{charge}_{a,z}$). The solution also provides dispatch variables for each time period $t$ yielding five output time series: 1) power output ($\pi_{a,t,z}$), 2) storage charge ($\Psi^{charge}_{a,t,z}$) 3) storage discharge ($\Psi^{discharge}_{a,t,z}$) 4) non-served energy ($\chi_{a,t,z})$ and 5) line power flow ($\phi_{t,z,z'}$).

To quantify the impact of input data variations on CEM outputs, we generate estimated outputs from solving the CEM for each single period in the underlying input data and use that information to inform selection of representation periods. If each representative period consists of $q$ hours, one year input data (8760 hours) will have $p = \lfloor \frac{8760}{q} \rfloor$  periods and correspondingly $p$ CEMs can be solved to generate output features. Note that the output produced for each period is an abstraction of the outputs obtained from solving the FCEM which considers 8760 hours of operations. The solution time required to solve the RCEM for each period can is generally much smaller than than solving the FCEM \cite{https://doi.org/10.48550/arxiv.2203.12426}. Moreover, the independence of CEM problems for each time periods allows multi-threading (or parallelization). From the outputs of the RCEM, we use the five output time series mentioned above as features to be used in the data encoding. 

\subsection{Dimensionality reduction - autoencoders}
We perform dimensionality reduction using a structure of autoencoders that refer to a type of neural network which enable transforming high-dimensional input data into their lower dimensional data (encoding) and vice-versa (decoding). The encoder transforms the high dimensional input data into lower dimension, referred to as \textit{latent representation}, while keeping the most important features. The decoder uses the latent representation of the data to reconstruct the initial input data.

\begin{figure}[!ht]
\centering
 \includegraphics[width=8cm, keepaspectratio]{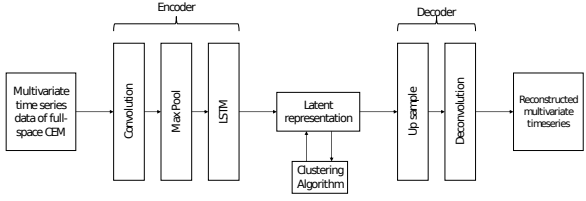}
  \caption{Autoencoder architecture used in the RPS method. Encoder is constructed using three main layers: Convolution, Maximum Pooling and, Long short term memory (LSTM) \cite{Goodfellow-et-al-2016}. Decoder is constructed using two main layers: Up sampling and deconvolution \cite{Goodfellow-et-al-2016}. Clustering is performed in the latent with the representative periods identified from clustering subsequently decoded to produce input time series for the RCEM.}
   \label{fig:arch}
\end{figure}

The encoder and decoder are constructed using various deep learning layers \cite{Goodfellow-et-al-2016} in Keras \cite{chollet2015keras} as follows: 

The encoder network architecture (Fig. \ref{fig:arch}) starts with a 1D convolution layer, which extracts the important short-term waveforms. The convolution layer is followed by a maximum pooling layer to reduce the dimension of the data by combining the output neurons of the convolution layer into a single neuron to the subsequent layer. Then a bidirectional long short-term memory layer is introduced to learn the temporal changes in both directions relative to each time period.  The decoder consists of a fully connected upsampling layer followed by convolutional transpose layer to transform the latent representation back to the original dimension. The objective of the autoencoder is to find a code for each data sample through minimization of the mean squared error (MSE) between its output and input over all samples. The parameters of the autoencoder are updated by minimization of the reconstruction mean squared error (MSE) as seen in Eq. \ref{eqn:ae_loss}:
\begin{equation}
\centering
    L_r = \frac{1}{p}\sum_{i=1}^p||D(E(x_i))-x_i||^2_2
    \label{eqn:ae_loss}
\end{equation}

where $p$ is the number of periods considered (e.g., $p = 52$ if the planning horizon is one year and each period is defined as a week --- $q = 168$ hours) in the data entries $x_i \in \mathbb{R}^{(n+l)\cdot q}$ is the $i^{th}$ period. $E$ and $D$ represent the encoder mapping (Eq. \ref{eqn:map1}) and decoder mapping (\ref{eqn:map2}), respectively, with $k$ being the desired number of clusters.

\begin{equation}
\centering
    E \colon \mathbb{R}^{(n+l)\cdot q} \mapsto \mathbb{R}^{\frac{(n+l)\cdot q}{k}}
    \label{eqn:map1}
\end{equation}

\begin{equation}
\centering
    D \colon \mathbb{R}^{\frac{(n+l)\cdot q}{k}} \mapsto \mathbb{R}^{(n+l)\cdot q}
    \label{eqn:map2}
\end{equation}

Moreover, we set the maximum pooling size to the number of clusters to reduce the inputted multivariate space to latent representation. Therefore, the data is being compressed from $\mathbb{R}^{p \times (n+l)\cdot q} $ to $\mathbb{R}^{p \times \frac{(n+l)\cdot q}{k}}$ where $k$ is the number of clusters desired. For autoencoders \cite{Goodfellow-et-al-2016}, when the dimension of the latent representation is on the order of the number of clusters of the input data then the network can be trained in an end-to-end manner without including regularization \cite{Guo2017} such as \textit{Dropout} and batch normalization terms \cite{JMLR:v15:srivastava14a, pmlr-v37-ioffe15}.

\subsection{Clustering algorithm}\label{clus_alg}
We use the unsupervised \textit{k}-means clustering algorithm that minimizes the sum of intra-cluster distances, with distance defined by the Euclidean norm in our case study. The objective function of the \textit{k}-means algorithm is given in Eq. \ref{eqn:c_loss} which is unsupervised. The objective function Eq. \ref{eqn:c_loss} of \textit{k}-means minimizes the mean squared error between the representative periods. In Eq. \ref{eqn:c_loss}, $S_j$ is set of periods assigned to cluster $j$, $k$ is the number of clusters desired, and $\bar{x}_j$ is the centroid of cluster $j$ \cite{MacKay2003}. For each cluster, we use the period that is nearest to the cluster centroid as the representative period \cite{MALLAPRAGADA20181231}. The algorithm is implemented using \textit{scikit-learn} library KMeans function with 10,000 repetitions of different centroid seeds to ensure that a global solution can be found \cite{scikit-learn}. The clustering algorithm produces period-indexed (52 in case of clustering of weeks) labels of the annual time series data. This labeling is used to generate the reduced time series data after the latent representation is reconstructed through the decoder.
\begin{equation}
\centering
    L_c = \sum_{j=1}^k\sum_{x_i \in S_j}||x_i-\bar{x}_j||^2_2
    \label{eqn:c_loss}
\end{equation}

\subsection{Overall loss function definition}
To guide the autoencoder to represent features that are important to forming clusters, the overall loss function for the RPS is defined by Eq. \ref{eqn:loss}. Here, $\gamma$ is a tunable parameter between 0 and 1 that adjusts the relative importance of the error induced in the latent representation and the error introduced via the clustering process.

\begin{equation}
\centering
    L = L_r + \gamma L_c
    \label{eqn:loss}
\end{equation}

\begin{figure}[!ht]
\centering
 \includegraphics[width=8cm, keepaspectratio]{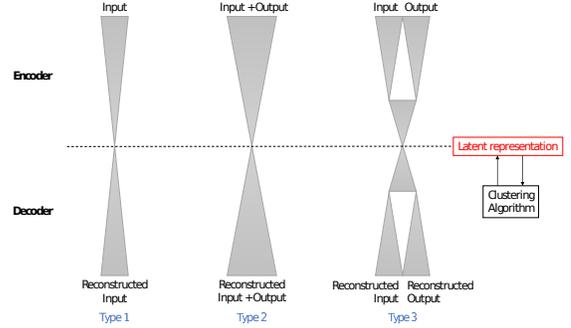}
  \caption{Three types autoencoder structures proposed. Type 1: input autoencoding only, Type 2: joint input and output autoencoding, Type 3: separate input and output autoencoding.}
   \label{fig:ae}
\end{figure}

We examine three alternative configurations for coupling autoencoders and the clustering process as illustrated in Fig. \ref{fig:ae}. Type 1 only uses input data, Type 2 uses both input and output data in one autoencoder and, Type 3 splits the autoencoding of the input and the output data into two separate autoencoders. Specifically, the outputs of both encoders in Type 3 are intermediate latent representations which are fed into a second-level autoencoder that produces the latent representation on which clustering is performed. The decoding side of the Type 3 method mirrors the encoding side. Both Type 1 and 2 autoencoders use the loss function defined by Eq. \ref{eqn:ae_loss}. Type 3 has an expanded autoencoder loss function as detailed in Eq. \ref{eqn:ae_type3_loss}, where $L_{int}$ is the intermediate autoencoder loss function, $L_I$ is the input data autoencoder loss function, $L_O$ is the output data autoencoder loss function and, $\alpha$ and $\beta$ are parameters that balance the feature importance of input relative to output data, while satisfying the condition: $\alpha +\beta = 1$. We identify the values of $\alpha$ and $\beta$ that minimize the overall loss function by scanning through 100 combinations of values that satisfy the above-mentioned condition. Note that $L_{r_i}$,$L_{I}$,$L_{O}$ are defined based on their respective input and output data using Eq. \ref{eqn:ae_loss}.

\begin{equation}
\centering
    L_r = L_{int}\times(\alpha L_I + \beta L_O)
    \label{eqn:ae_type3_loss}
\end{equation}

By jointly reducing input and output, the Type 2 autoencoder could in principle, identify input features that have a greater impacts on CEM outputs. However, even a dimensionality reduction technique will have diminishing returns with increasing number of features i.e. the dimension of data input to the encoder. This motivated us to consider the Type 3 encoder, where  we use a separate autoencoder for input and output and then combine them in a third autoencoder  (Eq. \ref{eqn:ae_type3_loss}).

Type 1 autoencoder only uses input data therefore it is comparable to \textit{k}-means on input data only (I \textit{k}-means). On the other hand, Type 2 and Type 3 autoencoders use both input and output data and therefore are comparable to \textit{k}-means on input and output data (I/O \textit{k}-means).

\subsection{Error metrics}

We focus on error metrics that quantify the difference between the results of the FCEM and the RCEM across multiple dimensions of installed capacity ($\Omega^{size}_{a,z}$,$\Omega^{energy}_{a,z}$,$\Omega^{charge}_{a,z}$), system cost (Eq. \ref{obj}), annual non-served energy (NSE, $\chi_{t,z}$ in Eqn. \ref{dem}, summed over the entire year), generation ($\pi_{t,z}$ in Eqn. \ref{dem}, summed over the entire year), while using the same time-independent parameters inputs. For capacity, we compare capacity results retrieved directly from the outputs of RCEM and FCEM. In RCEM, each resource $a$ in the set of resource $A$ and each zone $z$ in the set of zones $Z$ of a case $m$ ($x^{z,a,m}$) has a relative absolute error to corresponding FCEM ($y^{z,a,m}$) defined by the generalized Eq. \ref{error}. Where $m \in M$ corresponds to the set of cases defined by alternative technology assumptions, demand profiles, and representative periods (Appendix \ref{si} Table \ref{table:input}). We use absolute value based error metrics to account for both positive and negative deviations between the RCEM outputs and FCEM outputs.

\begin{equation}
    AE_{a,z,m} = \frac{|x_{a,z,m} - y_{a,z,m}|}{y_{a,z,m}}
    \label{error}
\end{equation}

Additionally, each case $m$ in $M$ has multiple generation sources $a \in A$ in zones $z \in Z$. We report the weighted average absolute error per case $m$ by the generalized equation Eq. \ref{wae_error}. The weight $y^{a,z,m}$ is the result of FCEM of the corresponding resource $a$ of zone $z$ in case $m$.

\begin{equation}
    WAE_{a,z,m} = \frac{\sum_{a \in A, z \in Z} AE_{a,z_m} y_{a,z,m}}{\sum_{a \in A, z \in Z}y_{a,z,m}}
    \label{wae_error}
\end{equation}

In the results, we report the mean and distribution of the weighted average absolute error of all the cases $m \in M$. The weighted average absolute capacity error is defined as the weighted average installed capacity error relative to the full-space results for each considered generator type (i.e., both energy generating and dependent resources). The mean capacity error metric is therefore defined by Eq. \ref{cap_error}, $AE^{\Omega^{size}}_{a,z,m}$, $AE^{\Omega^{energy}}_{a,z,m}$ and, $AE^{\Omega^{charge}}_{a,z,m}$ are the installed size, energy and charge capacity errors (individually defined by Eq. \ref{error}) of generator $a$ in zone $z$ from case $m$ and $\Omega_{a,z,m}^{size, FCEM}$, $\Omega_{a,m}^{energy, FCEM}$ and, $\Omega_{a,z,m}^{charge, FCEM}$ are the FCEM outcomes corresponding to installed size, energy and charge capacities for the same case $m$.

\begin{equation}
WAE^{Cap} = \frac{1}{|M|} \sum_{m\in M} \begin{bmatrix}

\frac{\sum_a \sum_z AE_{a,z,m}^{\Omega^{size}} \cdot \Omega_{a,z,m}^{size, FCEM}}{\sum_a \sum_z \Omega_{a,z,m}^{size, FCEM}} &\\

+\frac{\sum_a \sum_z AE_{a,z, m}^{\Omega^{energy}} \cdot \Omega_{a,z,m}^{energy, FCEM}}{\sum_a \sum_z \Omega_{a,z,m}^{energy, FCEM}}&\\

+\frac{\sum_a AE_{a,z,i}^{\Omega^{charge}} \cdot \Omega_{a,z, m}^{charge, FCEM}}{\sum_a \sum_z \Omega_{a,z,m}^{charge, FCEM}}\end{bmatrix}
\label{cap_error}
\end{equation}

For operational cost, generation and NSE results, direct comparison between the RCEM results and FCEM results is not plausible due to the different temporal resolutions used. Instead, we fix the capacity variable of the RCEM solution and evaluate the optimal dispatch for the fixed capacity over the entire year at an hourly resolution that can be compared against the dispatch results and cost of the FCEM. The mean system cost of electricity (SCOE) error is defined by Eq. \ref{scoe} which resembles the optimization model's objective function Eq. \ref{obj} without considering NSE ($\chi$) -  error in the NSE is considered separately. Eq. \ref{scoe_eqn} is the objective function of the CEM defined in Eq. \ref{obj} (noted as $Obj$) without the NSE.

\begin{equation} \label{scoe_eqn}
    SCOE_{a,z,m} = Obj_{a,z,m} - \sum_z \sum_t w_t \cdot \chi_{t, z} \cdot C^\chi_{z}
\end{equation}

We normalize $SCOE$ by the total demand to produce the error across all the cases $m$ with differing demand profiles as in Eq. \ref{scoe}, where $SCOE_{a,z,m}^{FCEM}$ is the full-space SCOE (as defined in Eq. \ref{scoe_eqn}) of generator $a$ in zone $z$.

\begin{eqnarray} \label{scoe}
\begin{split}
WAE^{SCOE} = \frac{1}{|M|} \sum_{m\in M} \frac{1}{\sum_t \sum_z L_{t,z,m}}\\
\qquad \times \frac{\sum_a \sum_z AE_{a,z,m}^{SCOE} \cdot SCOE_{a,z,m}^{FCEM}}{\sum_a \sum_z SCOE_{a,z,m}^{FCEM}}\\
\end{split}
\end{eqnarray}

NSE is reported as a percentage of the annual demand of the corresponding load year as described in Eq. \ref{nse}:

\begin{equation}
WAE^{NSE} = \frac{1}{|M|} \sum_{m\in M} \bigg(\frac{\sum_z \sum_t \chi_{t, z, m}}{\sum_t \sum_z L_{t,z,m}} \bigg)
\label{nse}
\end{equation}

The generation error is the generation weighted average error relative to the full-space for each considered generator type (i.e., energy generating resources only) as described in Eq. \ref{gen_error} where $AE^{\pi}_{a,m}$ an individual case's generation error (defined by Eq. \ref{error} of generator $a$ from case $m$ and $\pi_{a,t,z,m}^{FCEM}$ is the full-space generation output corresponding to the same case $m$. Note that all the error metrics used to evvaluate the performance of the various RPS methods are dimensionless by the definitions provided above.

\begin{equation}
WAE^{Gen} = \frac{1}{|M|} \sum_{m\in M} \frac{\sum_a AE_{a,m}^{\pi} \cdot \sum_t \pi_{a,t,z,m}^{FCEM}}{\sum_a \sum_t \pi_{a,t,z,m}^{FCEM}} 
\label{gen_error}
\end{equation}

\section{Case Study Description}\label{casestudy}

\begin{figure}[!ht]
\centering
 \includegraphics[width=6cm, keepaspectratio]{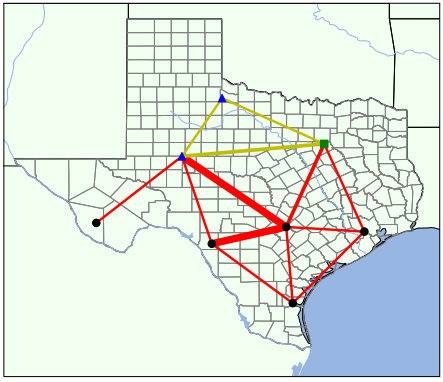}
  \caption{ERCOT electricity system. Single bus system (square green node); three bus network (square green node and blue triangle nodes); eight bus network is all inclusive. Line thickness highlights the average power transfer in 2018 between zones on a relative basis.}
   \label{ercot}
\end{figure}

To assess the performance of the autoencoder-based RPS method, we evaluate CEM results on three different case studies, all of which are based on the data and topology of the Electric Reliability Council of Texas (ERCOT). We consider representative period of length 168 hours with $k = $ 4, 8 and, 20 . Our experiments span three networks: single bus, three bus and eight bus systems as highlighted in Fig. \ref{ercot} and 90 scenario evaluations per representative period and network size pairing, where each scenario considers an alternative assumptions regarding time-independent parameters (e.g. VRE capital cost, see Appendix \ref{si} Table \ref{table:input}). The single bus system only considers the North Central region of Texas without network data. The three bus system consists of a complete linkage network of the North Central, South Central and Coastal regions of Texas. The eight bus systems considers all eight regions of Texas. Load data at the zonal level is provided by the ERCOT \cite{ercot_load}. We aggregate historical hourly solar and wind capacity factor data at the county level \cite{ercot_weather} to the zonal level according the ERCOT load zone map \cite{ercot_load}. Finally, we use a publicly available eight bus system \cite{ercot_network} to model the power flow and nodal connections of the ERCOT load zones as displayed in Fig. \ref{ercot}. Each load zone has four types of generation units: combined cycle gas turbine, solar, wind and short-duration battery storage. We use the National Renewable Energy Laboratory's Annual Technology Baseline \cite{ATB2020} for the relevant technical and cost parameters. 

At each load zone, the input time series data that must be clustered are: load, solar and wind profiles. Therefore, for the eight bus system  $168\times3\times8$ columns of length $52$ must be clustered into the desired number of representative weeks. The input time series has a shape of $52\times4,032$ with a latent representation of $52\times504$ given the autoencoder parameters (see Appendix \ref{si} Table \ref{table:param}) for the desired cluster size of $k = 8$. Since the pooling layer is set to the cluster size $k$, $504 = 4,032 \div 8$ is the latent representation size. While both learning and clustering performance can be tuned for better results on a case by case basis, we fix the hyperparameter values (see Appendix \ref{si} Table \ref{table:param}) of the autoencoders for all number of clusters and all cases so as to facilitate a consistent analysis. All autoencoders, clustering algorithms and optimization models were coded in the Python language and run on MIT Supercloud High Performance Computer using Intel Xeon Platinum 8260 processor with up to 48 cores and 192 GB of RAM \cite{reuther2018interactive}.

\section{Results}\label{results}
\subsection{CEM results comparison - single case}
\begin{figure}[!ht]
\centering
 \includegraphics[width=6cm, keepaspectratio]{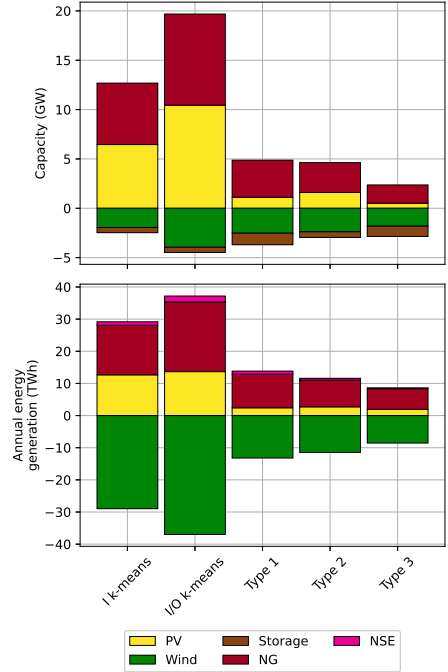}
 \caption{Difference in installed capacity and annual generation between reduced-space CEM (RCEM) results and full-space CEM results for three  bus network. Results based on RCEM using $k = 8$ for ERCOT load year 2020, mid-range VRE technology cost, 1,000 \$/tonne CO$_2$ price.}
    \label{res}
\end{figure}

Fig. \ref{res} highlights the impact of different RPS methods on installed capacities and annual generation outputs for the 3-bus case study and 8 representative weeks (results can be seen as a difference plot in Appendix \ref{si} Fig. \ref{diff}). The mean NSE of the three autoencoder-based RPS methods is 0.25 \% and the mean NSE for the I and I/O \textit{k}-means is 0.61\%. As compared to \textit{k}-means only based approaches, Type 1-3 methods also result in smaller deviations in annual generation results and capacity results as compared to the FCEM results (Fig. \ref{res}). Battery storage energy and charge capacities are relatively stable across all methods as seen in Fig. \ref{res} and Appendix \ref{si} Fig. \ref{diff}, with capacity errors below 3\% relative to the full-space results. The low battery storage capacity error can be attributed to the presence of two VRE sources (solar PV and wind) which diversify the dependency of charging and discharging on the availability of VRE generation. We note that both conventional RPS methods result in lesser wind resources and more solar relative to both the FCEM results and the RCEM outcomes using autoencoder based RPS. Wind availability profiles are more variable than solar, which is not adequately captured in the conventional RPS methods and leads to over investment in natural gas which is a dispatchable resource.

\subsection{CEM results comparison- multiple case}
The general nature of the results shown in Fig. \ref{res} for a single case are assessed by a broad set of numerical experiments over the three networks and four different values of representative periods. Table \ref{ev_res} summarizes the average value of different error metrics (over 90 parameter scenarios) related to system cost, NSE, annual generation and capacity (as per the definitions provided earlier see Eq. \ref{cap_error} - \ref{gen_error}) across the RPS methods and network case studies. We note that Type 2 and Type 3 autoencoders, which use estimated output data, noticeably improve all four error metrics relative to both \textit{k}-means and Type 1 autoencoder. In contrast, incorporating output data directly in the clustering process without dimensionality reduction, as in the I/O \textit{k}-means method, does not lead to improved error metrics as compared to the I \text{k}-means method. As pointed in \cite{https://doi.org/10.1002/sam.11161}, including more dimensions to an unsupervised learning algorithm, such as \textit{k}-means, may increase the complexity of identifying good quality solutions to the clustering problem.

\begin{figure*}[!ht]
\centering
 \includegraphics[width=12cm, keepaspectratio]{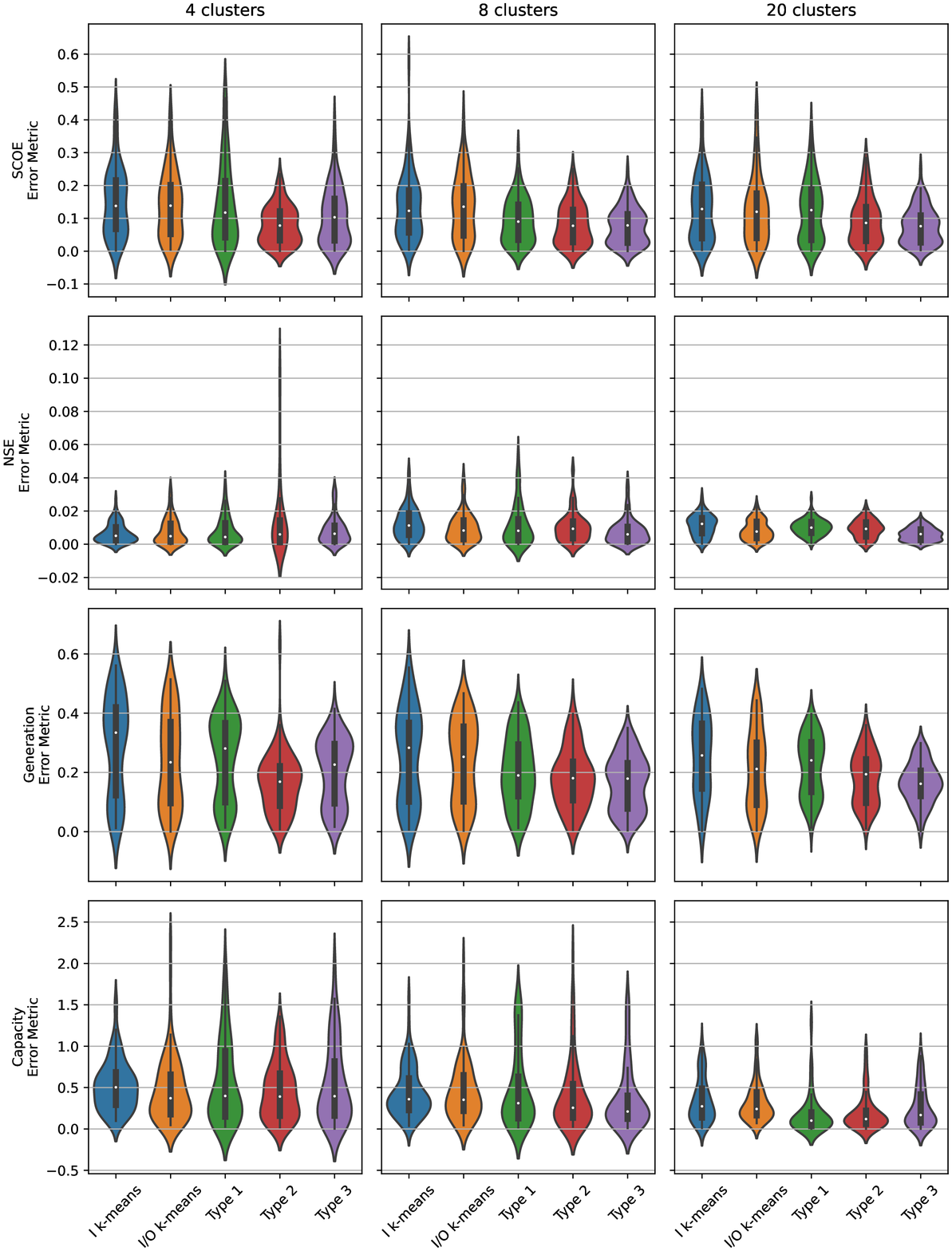}
  \caption{Violin plot featuring the kernel density estimator of the distribution of data points of the four weighted absolute error metrics (SCOE, NSE, generation, and capacity) grouped by RPS method for RCEM outcomes with different number of representative periods 4, 8, and 20. The data presented is based on the eight bus system discussed in section \ref{casestudy} and is consistent with the  mean values presented in Table \ref{scoe_res}.}
   \label{syscost}
\end{figure*}

\begin{table}[!ht]
\centering
  \caption{Mean values of error metric defined in Eq. \ref{cap_error} - \ref{gen_error} to quantify performance of different representative period selection (RPS) methods  for the 1-bus, 3-bus and 8-bus networks. Row-wise color mapping: red is highest and green is lowest value across the row.}
   \label{ev_res}
\begin{tabular}{c|c|cc|ccc}
\hline
 & \textbf{bus} & \textbf{I \textit{k}-means} & \textbf{I/O \textit{k}-means} & \textbf{Type 1} & \textbf{Type 2} & \textbf{Type 3} \\
\hline
\hline
 & 1 & \cellcolor[HTML]{F8696B}12.3 & \cellcolor[HTML]{FB9273}11.97 & \cellcolor[HTML]{FFEB84}11.23 & \cellcolor[HTML]{C5DA80}9.97 & \cellcolor[HTML]{63BE7B}7.8 \\
 & 3 & \cellcolor[HTML]{F8696B}12.79 & \cellcolor[HTML]{FA8E72}12.64 & \cellcolor[HTML]{FFEB84}12.25 & \cellcolor[HTML]{9BCE7E}9.45 & \cellcolor[HTML]{63BE7B}7.85 \\
\multirow{-3}{*}{\rotatebox[origin=c]{90}{SCOE}} & 8 & \cellcolor[HTML]{F8696B}14.24 & \cellcolor[HTML]{F9756E}14.05 & \cellcolor[HTML]{FFEB84}12.07 & \cellcolor[HTML]{63BE7B}8.57 & \cellcolor[HTML]{7AC47C}9.09 \\
\hline
 & 1 & \cellcolor[HTML]{F8696B}1.12 & \cellcolor[HTML]{DFE182}0.85 & \cellcolor[HTML]{F9706D}1.11 & \cellcolor[HTML]{FFEB84}0.91 & \cellcolor[HTML]{63BE7B}0.61 \\
 & 3 & \cellcolor[HTML]{F8696B}1.16 & \cellcolor[HTML]{FFEB84}0.96 & \cellcolor[HTML]{FDC47D}1.02 & \cellcolor[HTML]{E7E482}0.92 & \cellcolor[HTML]{63BE7B}0.7 \\
\multirow{-3}{*}{\rotatebox[origin=c]{90}{NSE}} & 8 & \cellcolor[HTML]{F8696B}1.33 & \cellcolor[HTML]{FFEB84}0.99 & \cellcolor[HTML]{FFE082}1.02 & \cellcolor[HTML]{9ECF7E}0.85 & \cellcolor[HTML]{63BE7B}0.77 \\
\hline
 & 1 & \cellcolor[HTML]{F8696B}23.9 & \cellcolor[HTML]{FFEB84}19.12 & \cellcolor[HTML]{FCAD78}21.43 & \cellcolor[HTML]{DFE182}18.3 & \cellcolor[HTML]{63BE7B}15.1 \\
 & 3 & \cellcolor[HTML]{F8696B}25.05 & \cellcolor[HTML]{FFEB84}20.96 & \cellcolor[HTML]{FED07F}21.84 & \cellcolor[HTML]{99CD7E}17.94 & \cellcolor[HTML]{63BE7B}16.32 \\
\multirow{-3}{*}{\rotatebox[origin=c]{90}{Gen}} & 8 & \cellcolor[HTML]{F8696B}27.13 & \cellcolor[HTML]{FECE7F}23.35 & \cellcolor[HTML]{FFEB84}22.24 & \cellcolor[HTML]{63BE7B}17.47 & \cellcolor[HTML]{6CC07B}17.77 \\
\hline
 & 1 & \cellcolor[HTML]{F8696B}29.53 & \cellcolor[HTML]{FA8471}27.46 & \cellcolor[HTML]{B2D57F}17.54 & \cellcolor[HTML]{63BE7B}15.73 & \cellcolor[HTML]{FFEB84}19.27 \\
 & 3 & \cellcolor[HTML]{F8696B}34.76 & \cellcolor[HTML]{FB9674}31.8 & \cellcolor[HTML]{70C17B}20.11 & \cellcolor[HTML]{FFEB84}26.1 & \cellcolor[HTML]{63BE7B}19.56 \\
\multirow{-3}{*}{\rotatebox[origin=c]{90}{Cap}} & 8 & \cellcolor[HTML]{FDBE7C}44.22 & \cellcolor[HTML]{FFEB84}42.61 & \cellcolor[HTML]{F8696B}47.19 & \cellcolor[HTML]{B6D67F}38.61 & \cellcolor[HTML]{63BE7B}34.00 \\
 \hline
\end{tabular}
\end{table}
To explore the scalability of the RPS methods, we investigated the impact of different representative period choices on the various error metrics using the 8-bus network as the case study (Table \ref{scoe_res} and Fig. \ref{syscost}). When comparing the mean value of the error metrics of NSE, system cost and annual generation, Table \ref{scoe_res} highlights that Type 3 autoencoder consistently outperforms the other methods across all cluster numbers. In addition to the mean values, Fig. \ref{syscost} shows that the distribution of error metrics across the 90 experiments carried out are narrower for the Type 3 encoder as compared to the other RPS methods. In particular, for the Type 3 autoencoder, the probability density near zero is larger as compared to other RPS methods. For a total of 90 distinct cases for each cluster size $k$ (4, 8, and 20) per method in Fig. \ref{fig:ae}, the Type 3 autoencoder is able to satisfy 74\% of the cases' peak demand according to the dispatch-only optimization as compared to 63\% for the I/O \textit{k}-means method, 70\% for the Type 1 autoencoder method and 62\% for the I \textit{k}-means method. Furthermore the standard deviation in the capacity error metric for the eight bus system (see Appendix \ref{si} Table \ref{table:stdev}) not only reflects the performance of the autoencoder-based RPS but also the consistency. Note that the violin plots in Fig. \ref{syscost} extrapolate data points to visualize the kernel density estimator which may produce negative values, however the box plot portion of the violin plot (i.e., data points) are all greater than zero values since absolute errors are reported.

\begin{table}[!ht]
\centering
  \caption{Mean error metric results of different representative period selection (RPS) methods grouped by cluster number $k$ for the 8-bus system. Row-wise color mapping: red is highest and green is lowest value across the row.}
   \label{scoe_res}
\begin{tabular}{c|c|cc|ccc}
\hline
 & \textbf{$k$} & \textbf{I \textit{k}-means} & \textbf{I/O \textit{k}-means} & \textbf{Type 1} & \textbf{Type 2} & \textbf{Type 3} \\
\hline
\hline
 & 4 & \cellcolor[HTML]{F8696B}14.98 & \cellcolor[HTML]{FFDB81}14.30 & \cellcolor[HTML]{FFEB84}14.20 & \cellcolor[HTML]{63BE7B}7.94 & \cellcolor[HTML]{AAD27F}10.80 \\
 & 8 & \cellcolor[HTML]{F96F6D}13.47 & \cellcolor[HTML]{F8696B}13.65 & \cellcolor[HTML]{FFEB84}9.24 & \cellcolor[HTML]{A7D17E}8.33 & \cellcolor[HTML]{63BE7B}7.61 \\
\multirow{-3}{*}{\rotatebox[origin=c]{90}{SCOE}} & 20 & \cellcolor[HTML]{F8696B}14.27 & \cellcolor[HTML]{F96F6C}14.21 & \cellcolor[HTML]{FFEB84}12.77 & \cellcolor[HTML]{7AC47C}9.44 & \cellcolor[HTML]{63BE7B}8.85 \\
\hline
 & 4 & \cellcolor[HTML]{F8696B}1.50 & \cellcolor[HTML]{E6E482}0.78 & \cellcolor[HTML]{FFEA84}0.81 & \cellcolor[HTML]{63BE7B}0.67 & \cellcolor[HTML]{FFEB84}0.80 \\
 & 8 & \cellcolor[HTML]{F8696B}1.28 & \cellcolor[HTML]{F8E983}1.00 & \cellcolor[HTML]{FB9E76}1.17 & \cellcolor[HTML]{FFEB84}1.01 & \cellcolor[HTML]{63BE7B}0.76 \\
\multirow{-3}{*}{\rotatebox[origin=c]{90}{NSE}} & 20 &\cellcolor[HTML]{F8696B}1.22 & \cellcolor[HTML]{FA8571}1.19 & \cellcolor[HTML]{FFEB84}1.08 & \cellcolor[HTML]{A0CF7E}0.88 & \cellcolor[HTML]{63BE7B}0.75 \\
\hline
 & 4 & \cellcolor[HTML]{F8696B}28.89 & \cellcolor[HTML]{FFEB84}23.83 & \cellcolor[HTML]{FED280}24.83 & \cellcolor[HTML]{63BE7B}16.38 & \cellcolor[HTML]{B2D47F}20.18 \\
 & 8 & \cellcolor[HTML]{F8696B}25.45 & \cellcolor[HTML]{FB9774}23.57 & \cellcolor[HTML]{FFEB84}20.10 & \cellcolor[HTML]{ADD37F}18.13 & \cellcolor[HTML]{63BE7B}16.32 \\
\multirow{-3}{*}{\rotatebox[origin=c]{90}{Gen}} & 20 &\cellcolor[HTML]{F8696B}27.04 & \cellcolor[HTML]{FED680}22.66 & \cellcolor[HTML]{FFEB84}21.79 & \cellcolor[HTML]{85C77C}17.89 & \cellcolor[HTML]{63BE7B}16.80 \\
\hline
 & 4 & \cellcolor[HTML]{FCA477}52.12 & \cellcolor[HTML]{FFEB84}46.89 & \cellcolor[HTML]{F8696B}56.34 & \cellcolor[HTML]{F3E783}46.40 & \cellcolor[HTML]{63BE7B}40.07 \\
 & 8 & \cellcolor[HTML]{FFEB84}44.03 & \cellcolor[HTML]{FDC37D}45.31 & \cellcolor[HTML]{F8696B}48.10 & \cellcolor[HTML]{B5D57F}40.18 & \cellcolor[HTML]{63BE7B}35.83 \\
\multirow{-3}{*}{\rotatebox[origin=c]{90}{Cap}} & 20 & \cellcolor[HTML]{FBA076}36.50 & \cellcolor[HTML]{FFEB84}35.62 & \cellcolor[HTML]{F8696B}37.13 & \cellcolor[HTML]{96CC7D}29.26 & \cellcolor[HTML]{63BE7B}26.10 \\
 \hline
\end{tabular}
\end{table}

\subsection{Impact of input data on CEM results}
As previously mentioned, the Type 3 RPS method expands the loss function to Eq. \ref{eqn:ae_type3_loss}, where $L_I$ is the input data autoencoder loss function and $L_O$ is the output data autoencoder loss function given $\alpha$ and $\beta$ as tunable parameters that balance the feature importance of input relative to output data. The tunable parameters satisfy $\alpha +\beta = 1$. In systems with less variable time series patterns such as a solar only VRE system without storage, the CEM outputs can be predicted quite well with input data and hence one would expect $\beta$ to be relatively small compared $\alpha$. However, in systems with more variability in grid operations, such as a combination of VRE and battery storage, makes the system dispatch more complicated as VRE are volatile resources and storage follows a complex charge/discharge pattern. In this case, information about estimated outputs are likely to be more important than the solar only case, hence leading to a higher value of $\beta$. This hypotheses can be evaluated by exploring the optimal value of the input and output feature importance parameters $\alpha$ and $\beta$ for the Type 3 autoencoder in the case of CEM applied to simplified case studies. 

\begin{figure}[!ht]
\centering
 \includegraphics[width=9cm, keepaspectratio]{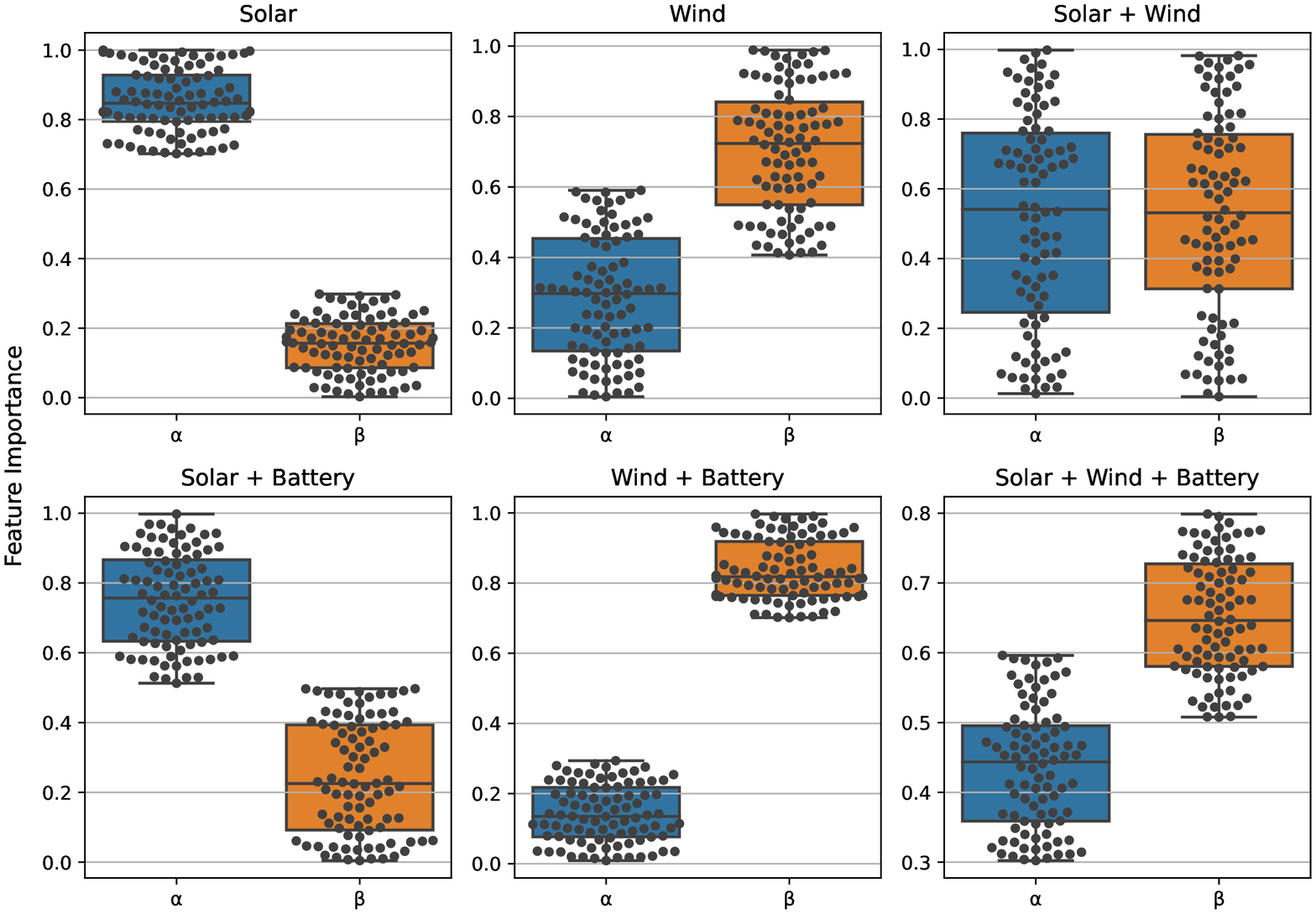}
  \caption{Box and swarm plot of optimized values of $\alpha$ and $\beta$ parameters for Type 3 autoencoder across six simplified case studies described in section \ref{clus_alg}. Results based on RCEM outcomes for single bus system with 4, 8, and 20 representative periods for all 90 distinct cases as defined by parameter values in Appendix \ref{si} Table \ref{table:input}.}
   \label{fig:alphabeta}
\end{figure}

Figure \ref{fig:alphabeta} summarizes the values of $\alpha$ and $\beta$ for six 1-bus systems where the resources are restricted to the following options: 1) natural gas and solar, 2) natural gas, solar and battery, 3) natural gas and wind, 4) natural gas, wind and battery, 5) natural gas, solar and wind, 6) natural gas, solar, wind and battery.  We perform this sensitivity analysis on 1-bus case to eliminate power flow and network effects and highlight the impact of input and output time series on the behavior of the Type 3 autoencoder. Figure \ref{fig:alphabeta} shows that $\beta$ (output data importance) is higher when more time series variability (i.e., wind) is considered in the generation design. Furthermore, $\beta$ increases when battery storage is considered in the generation mix. This is due to the addition of the approximate charging and discharging output data that will improve the autoencoder's performance in identifying important features that can reproduce the FCEM results using the RCEM.

\subsection{Solution time comparison}
\begin{figure}[!ht]
\centering
 \includegraphics[width=6cm, keepaspectratio]{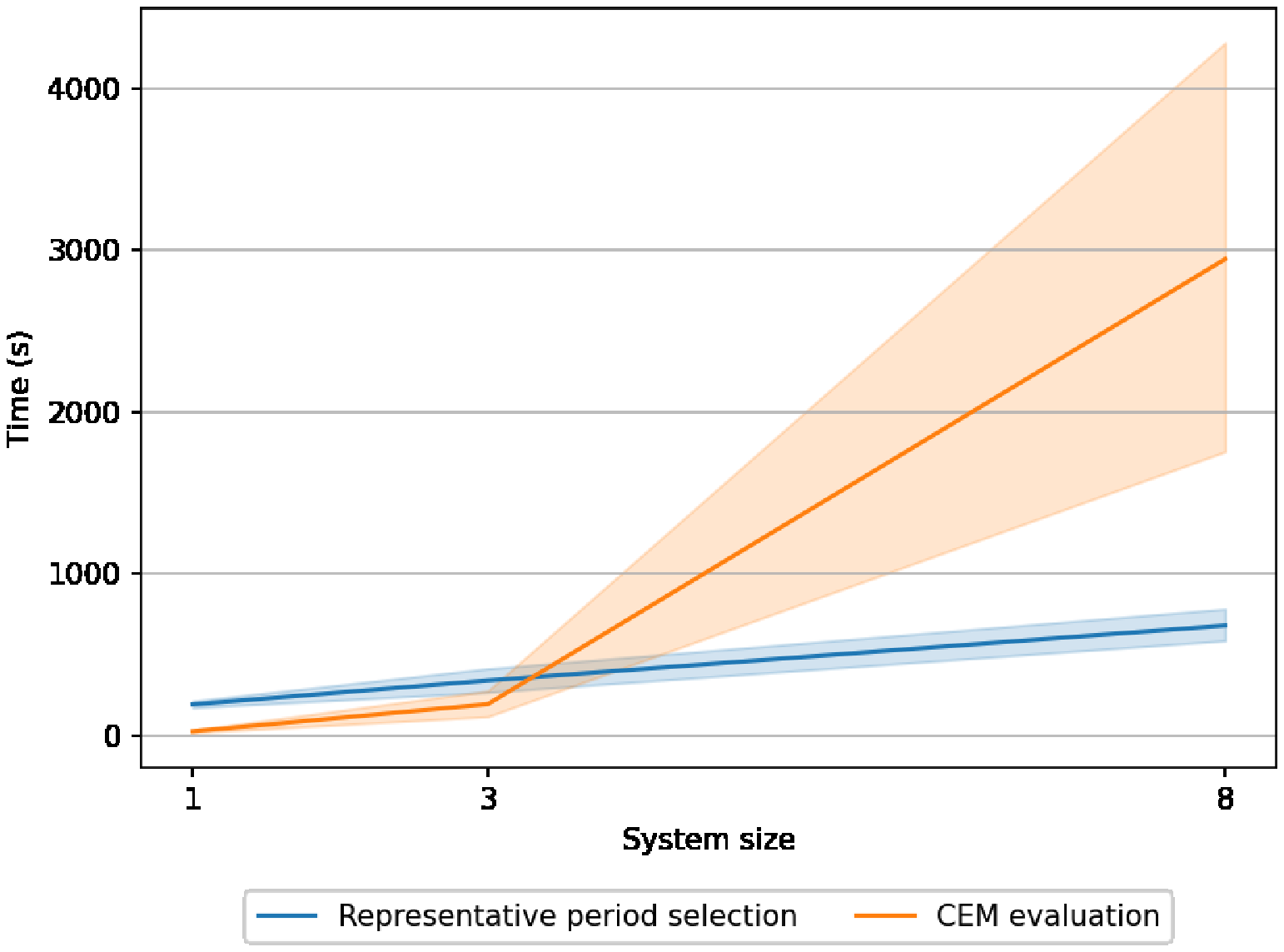}
  \caption{Mean computational time and 95\% confidence band of RCEM evaluation and all autoencoder-based representative period selection methods (Type 1, 2, and 3) for all sizes and scenarios across the single, three and eight bus systems.}
   \label{time}
\end{figure}

Evidently, the time to execute the autoencoder-based RPS methods is significantly longer than the standalone RPS methods. Appendix \ref{si} Fig. \ref{time} shows that run time RPS method may be larger than CEM solution time in the case of small systems such as the 1-bus cases and some of the 3-bus cases. However, for larger systems i.e., more realistic CEM use cases, the autoencoder-based RPS method run time is dwarfed by the optimization run time for any cluster number $k$ (see Fig. \ref{time}).

\section{Conclusion} \label{conclusion}
In this paper, we present an autoencoder-based RPS method to enable computationally efficient solution of CEMs for large-scale power systems without sacrificing accuracy. We show that autoencoders can be better trained when considering both the input time series of an optimization model and an expected output signal that is obtained from CEM evaluation of disjoint time periods in the data set. The expected output signal need not be accurate but sufficiently directs the RPS method with the objective of translating the data to learn optimization model behavior. This is particularly important for optimization models with inter-temporal dependencies.

We report a rigorous statistical analysis of the proposed methods. The proposed Type 3 autoencoder generally outperforms all other methods on error metrics related to cost, NSE, annual generation and capacity. Furthermore, Type 3 autoencoder RPS of 4 and 8 representative weeks performs better than all other methods clustered at 8 and 20 representative weeks. Therefore the proposed architecture of the Type 3 autoencoder can significantly reduce the temporal resolution of grid operations to be modeled within a CEM without sacrificing accuracy of results or incurring significant run time burden. 

In summary, we present a statistical analysis that supports the use of autoencoders with clustering algorithms to reduce the error metric in electricity resource capacity expansion planning. Further parameter tuning and loss function modification can improve error metric results depending on the objective function and the clustering algorithm of choice. Here, we tested the proposed RPS method on deterministic CEM formulations with a single investment stage. However, these methods could be more impactful in enabling computationally efficient solution of stochastic, multi-stage electricity resource CEM problems as well as those considering coordinated investment planning across multiple infrastructures.

\section{Code and data availability} \label{code}

The capacity expansion model and autoencoder architecture codes are available at \cite{optmodel, autoencoder}. Additionally, the results of the  statistical analysis are available at \cite{marc_barbar_2022_6484347}.

\section{Acknowledgements}
This research was supported by MIT Energy Initiative's Future Energy System Center. The authors thank Prof. Robert Stoner and Rahman Khorramfar for their feedback on the manuscript. The authors acknowledge the MIT SuperCloud and Lincoln Laboratory Supercomputing Center for providing HPC resources that have contributed to the research results reported within this paper.

\appendix\label{si}

\begin{figure}[!ht]
\centering
 \includegraphics[width=8cm, keepaspectratio]{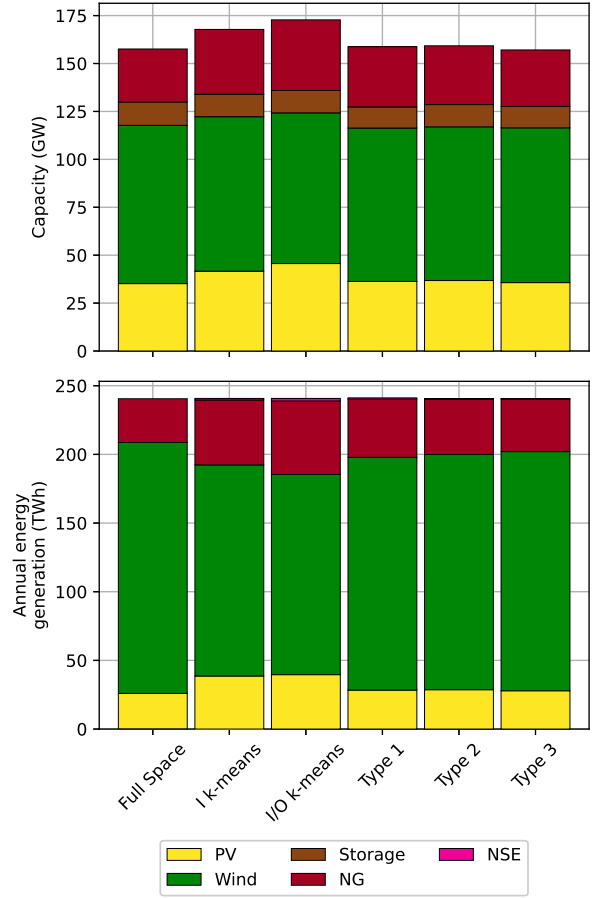}
   \caption{Installed capacity (top) and annual generation (bottom) for reduced-space CEM (RCEM) results and full-space CEM results for three  bus network. Results based on RCEM using $k = 8$ for ERCOT load year 2020, mid-range VRE technology cost, 1,000 \$/tonne CO$_2$ price.}
    \label{diff}
\end{figure}

\begin{table}[!ht]
\centering
  \caption{Input data considered in statistical analysis. Complete description of each parametric scenario, including description of low, medium and high technology cost assumptions are defined elsewhere \cite{marc_barbar_2022_6484347}.}
   \label{table:input}
\begin{tabular}{c|c|c|c}
\hline
Solar Tracking & Load Year& Technology Cost & CO$_2$ \$/tonne \\
\hline
\hline
Single & 2018 & Low & 50 \\
Dual & 2019 & Mid & 100 \\
 & 2020 & High & 500 \\
 &  &  & 1,000 \\
 &  &  & 1,500 \\
 \hline
\end{tabular}
\end{table}

\begin{table}[!ht]
\centering
  \caption{Autoencoder parameters. Encoder maximum pooling layer size is divisible by flattened time series of one period to satisfy dimensionality reduction to cluster latent representation by $k$.}
   \label{table:param}
\begin{tabular}{c|c}
\hline
Parameter & Value \\
\hline
\hline
Filter size & 50 \\
Kernel size & 10 \\
Stride & 1 \\
\hline
Pooling & Cluster size ($k$) \\
\hline
Long-short term memory (LSTM) Units & 50 \\
\hline
Deconvolution size & 10 \\
 \hline
\end{tabular}
\end{table}

\begin{table}[!ht]
\centering
  \caption{Standard deviation of capacity error metric grouped by cluster number $k$ for the eight bus system. Row-wise color mapping: red is highest and green is lowest.}
   \label{table:stdev}
\begin{tabular}{c|ccccc}
\hline
 & \textbf{I \textit{k}-means} & \textbf{I/O \textit{k}-means} & \textbf{Type 1} & \textbf{Type 2} & \textbf{Type 3} \\
 \hline
 \hline
4 & \cellcolor[HTML]{63BE7B}30.24 & \cellcolor[HTML]{FED580}39.35 & \cellcolor[HTML]{F8696B}46.33 & \cellcolor[HTML]{B5D57F}34.27 & \cellcolor[HTML]{FFEB84}37.87 \\
8 & \cellcolor[HTML]{F9E983}29.3 & \cellcolor[HTML]{F8696B}35.83 & \cellcolor[HTML]{FFEB84}29.39 & \cellcolor[HTML]{FDC67D}31.25 & \cellcolor[HTML]{63BE7B}26.69 \\
20 & \cellcolor[HTML]{FDBB7B}27.86 & \cellcolor[HTML]{CADB80}26.1 & \cellcolor[HTML]{F8696B}29.64 & \cellcolor[HTML]{FFEB84}26.8 & \cellcolor[HTML]{63BE7B}24.72 \\
\hline
\end{tabular}
\end{table}

\subsection{Capacity expansion model formulation}\label{optmodel}
The below model's nomenclature is presented in Table \ref{table:nomencl}.

\begin{eqnarray} \label{obj}
\begin{split}
\min & \sum_{a}\sum_{z}\big(\Omega^{size}_{a,z}\cdot (C^I_{a,z}+C^F_{a,z})\\
    &\qquad \qquad +\Omega^{energy}_{a,z}\cdot C^e_{a,z} \cdot (1+C^d_{a,z})\\\\
    &\qquad \qquad +\Omega^{charge}_{a,z}\cdot (C^c_{a,z}+C^{Fc}_{a,z})\\
    &\qquad \qquad +\sum_t w_t \cdot \pi_{a,t,z} \cdot (C^V_{a,z}+C^{Vf}_{a,z})\\
    &\qquad \qquad +\sum_t w_t \cdot \Psi^{charge}_{a,t,z} \cdot C^{Vc}_{a,z} \\
    &\qquad \qquad +\sum_t w_t \cdot n_{a,t,z} \cdot C^{start}_{a,z}\big)\\
    &+ \sum_z \sum_t w_t \cdot \chi_{t, z} \cdot C^\chi_{z} \\
\end{split}
\end{eqnarray}

\begin{eqnarray} \label{dem}
\begin{split}
\textrm{s.t.}& \quad L_{t,  z} = \sum_a(\pi_{a, t,  z}  + \Psi^{discharge}_{a,t,z}\\
&\qquad -\Psi^{charge}_{a,t,z}) + \chi_{t,  z}  + \phi_{t,z,z'}\\
&\forall z, z'\in Z \quad \forall t \in T
\end{split}
\end{eqnarray}

\begin{eqnarray} \label{vre}
\begin{split}
\pi_{a,t,z} \leq \Omega^{size}_{a,z} \cdot A_{a,t,z}  \\ 
\forall a \in R, \forall t \in T, \forall z \in Z
\end{split}
\end{eqnarray}

\begin{eqnarray}\label{stor1}
\begin{split}
    &\quad \Gamma_{a,t,z} = \Gamma_{a,t-1,z} - \frac{\psi^{discharge}_{a,t,z}}{\eta^{discharge}_{a,z}} + \eta^{charge}_{a,z} \cdot \psi^{charge}_{a, t,z}\\
    &\quad \forall a \in S,\forall t \in T^{interior} \in T, \forall z \in Z
\end{split}
\end{eqnarray}

\begin{eqnarray}\label{stor1_1}
\begin{split}
    &\quad \Gamma_{a,t,z} = \Gamma_{a,t^{period},z} - \frac{\psi^{discharge}_{a,t,z}}{\eta^{discharge}_{a,z}} + \eta^{charge}_{a,z} \cdot \Psi^{charge}_{a, t,z}\\
    &\quad \forall a \in S, \forall t \in T^{start} , \forall z \in Z
\end{split}
\end{eqnarray}

\begin{eqnarray}\label{stor5}
\begin{split}
\Gamma_{a,t,z} \leq \delta_{a,z} \cdot \Omega^{energy}_{a,z} \\ \forall a \in S, \forall t \in T, \forall z \in Z
\end{split}
\end{eqnarray}

\begin{eqnarray}\label{stor3}
\begin{split}
\Psi_{a,t,z}^{charge} \leq \Omega^{charge}_{a,z} \\ \forall a \in S, \forall t \in T, \forall z \in Z
\end{split}
\end{eqnarray}

\begin{eqnarray}\label{stor31}
\begin{split}
\Psi_{a,t,z}^{charge} + \Psi_{a,t,z}^{discharge} \leq \Omega^{charge}_{a,z} \\ \forall a \in S, \forall t \in T, \forall z \in Z
\end{split}
\end{eqnarray}

\begin{eqnarray}\label{stor4}
\begin{split}
\Psi_{a,t,z} \leq \Gamma_{a,t-1,z} \\ \forall a \in S, \forall t \in T, \forall z \in Z
\end{split}
\end{eqnarray}

\begin{eqnarray}\label{therm1}
\begin{split}
v_{a,t,z} \leq \frac{\Omega^{size}_{a,z}}{\Omega^{unit}_{a,z}} \\ \forall a \in M, \forall t \in T, \forall z \in Z
\end{split}
\end{eqnarray}

\begin{eqnarray}\label{therm11}
\begin{split}
u_{a,t,z} \leq \frac{\Omega^{size}_{a,z}}{\Omega^{unit}_{a,z}} \\ \forall a \in M, \forall t \in T, \forall z \in Z
\end{split}
\end{eqnarray}

\begin{eqnarray}\label{therm12}
\begin{split}
n_{a,t,z} \leq \frac{\Omega^{size}_{a,z}}{\Omega^{unit}_{a,z}} \\ \forall a \in M, \forall t \in T, \forall z \in Z
\end{split}
\end{eqnarray}

\begin{eqnarray}\label{therm1_1}
\begin{split}
v_{a,t,z} = v_{a,t-1,z} + u_{a,t,z} - n_{a,t,z} \\ \forall a \in M, \forall t \in T^{interior}, \forall z \in Z
\end{split}
\end{eqnarray}

\begin{eqnarray}\label{therm1_2}
\begin{split}
v_{a,t,z} = v_{a,t+t^{period}-1,z} + u_{a,t,z} - n_{a,t,z}\\ \forall a \in M, \forall t \in T^{start}, \forall z \in Z
\end{split}
\end{eqnarray}

\begin{eqnarray}\label{therm2}
\begin{split}
\pi_{a,t,z} \geq \rho^{min}_{a,z} \cdot \Omega^{size}_{a,z} \cdot v_{a,t,z} \\ \forall a \in M, \forall t \in T, \forall z \in Z
\end{split}
\end{eqnarray}

\begin{eqnarray}\label{therm3}
\begin{split}
\pi_{a,t,z} \leq \rho^{max}_{a,z} \cdot \Omega^{size}_{a,z} \cdot v_{a,t,z} \\ \forall a \in M, \forall t \in T, \forall z \in Z
\end{split}
\end{eqnarray}

\begin{eqnarray}\label{therm4}
\begin{split}
\pi_{a,t,z} - \pi_{a,t-1,z} \leq \Omega^{size}_{a,z} \cdot \kappa_{a,z}^{up} \\ \forall a \in M, \forall t \in T, \forall z \in Z
\end{split}
\end{eqnarray}

\begin{eqnarray}\label{therm5}
\begin{split}
\pi_{a,t-1,z} - \pi_{a,t,z} \leq \Omega^{size}_{a,z} \cdot \kappa_{a,z}^{down} \\ \forall a \in M, \forall t \in T, \forall z \in Z
\end{split}
\end{eqnarray}

\begin{eqnarray}\label{net1}
\begin{split}
\phi_{t,z,z'} =  B_{z,z'} \cdot (\theta_{t,  z}-\theta_{t,z'}) \\ \forall z, z' \in Z, \forall t \in T
\end{split}
\end{eqnarray}

\begin{eqnarray}\label{net2}
\begin{split}
\phi_{t,z,z'} \leq \Phi^{max}_{z,z'}
\\ \forall i, j \in Z, \forall t \in T
\end{split}
\end{eqnarray}

\begin{eqnarray}\label{net3}
\begin{split}
\phi_{t,z,z'} \geq -\Phi^{max}_{z,z'}
\\ \forall i, j \in Z, \forall t \in T
\end{split}
\end{eqnarray}

\begin{eqnarray}\label{nnc}
\begin{split}
\Omega^{charge}_{a,z}, \Omega^{discharge}_{a,z}, \Omega^{size}_{a,z} \geq 0 \\
\pi_{a,t,z}, \chi_{t,z}, \Gamma_{a,t,z}  \geq 0 \\
\Psi^{charge}_{a,t,z}, \Psi^{discharge}_{a,t,z} \geq 0\\
\theta^{min}_{t,z,z'} \leq \theta_{t,  z}-\theta_{t,z'} \leq \theta^{max}_{t,z,z'} \\
\theta_{t,  1} = 0\\
\phi_{t,z,z'} \in \mathbb{R}\\
\forall z,z' \in Z, \forall t \in T, \forall a \in M \cup R \cup S
\end{split}
\end{eqnarray}

\begin{table}[!ht]
\centering
\caption{Nomenclature of the electricity resource capacity expansion model of Appendix \ref{optmodel}}
\begin{tabular}{|c|c|}
\hline
\textbf{Set} & \textbf{Description} \\
\hline
$R$&Variable renewable energy resources\\
$S$&Battery storage resources\\
$M$&Thermal generation resources\\
$Z$&Power system zones\\
$T$&Hours of operation in a model period\\
$T^{interior}$&Interior time steps inside T\\
$T^{start}$&First time step of T\\
\hline
\textbf{Index} & \textbf{Description} \\
\hline
$a$&Generation resource\\
$t$&Time step\\
$z, z'$&Load zone\\
\hline
\textbf{Parameter} & \textbf{Description} \\
\hline
$t^{period}$&Total number of time steps\\
$w_t$&Time step weight\\
$C^I$&Investment cost (USD/MW)\\
$C^e$&Energy capacity investment cost (USD/MWh)\\
$C^c$&Charge investment cost (USD/MWh)\\
$C^d$&Battery energy capacity degradation per annum (\%)\\
$C^F$&Fixed operational cost (USD/MW-yr)\\
$C^{Fc}$&Fixed operational charge cost (USD/MWh-yr)\\
$C^V$&Variable cost (USD/MWh)\\
$C^{Vf}$&Fuel cost (USD/MWh)\\
$C^{Vc}$&Variable charge cost (USD/MWh)\\
$C^{\chi}$&Value of lost load (USD/MWh)\\
$C^{start}$&Startup cost (USD/MW)\\
$\mu$&Storage round-trip efficiency\\
$A$&Generation availability profile\\
$\Omega^{max}$&Maximum generation capacity\\
$\rho^{min}$&Minimum generation power\\
$\rho^{max}$&Maximum generation power\\
$\kappa^{up}$&Ramp up limit\\
$\kappa^{down}$&Ramp down limit\\
$B$&Line susceptance\\
$\Phi^{max}$&Maximum line power capacity\\
$\delta$&Storage depth of discharge\\
$\eta$&Storage efficiency\\
$\Omega^{unit}$&Generation unit capacity\\
$\theta^{min}, \theta^{max}$&Minimum and maximum voltage angle\\
$\Phi^{max}_{z,z'}$ & Maximum line power\\

\hline
\textbf{Variable} & \textbf{Description} \\
\hline

$\Omega^{size}_{a,z}$&Generation capacity\\
$\Omega^{energy}_{a,z}$&Energy capacity\\
$\Omega^{charge}_{a,z}$&Charge capacity\\
$\pi_{a,t,z}$&Power output\\
$v_{a,t,z}$&Number of units committed\\
$u_{a,t,z}$&Number of startup decisions\\
$n_{a,t,z}$&Number of shutdown decisions\\
$\Psi^{charge}_{a,t,z}$&Storage charge\\
$\Psi^{discharge}_{a,t,z}$&Storage discharge\\
$\chi_{t,z}$&Non-served energy\\
$\Gamma_{a,t,z}$&Storage state of charge\\
$\theta_{t,z}$&Line voltage angle\\
$\phi_{t,z,z'}$&Line power flow\\
\hline
\end{tabular}
\label{table:nomencl}
\end{table}

\ifCLASSOPTIONcaptionsoff
  \newpage
\fi

\bibliographystyle{IEEEtran}
\bibliography{main}

\begin{thebibliography}{10}
\providecommand{\url}[1]{#1}
\csname url@samestyle\endcsname
\providecommand{\newblock}{\relax}
\providecommand{\bibinfo}[2]{#2}
\providecommand{\BIBentrySTDinterwordspacing}{\spaceskip=0pt\relax}
\providecommand{\BIBentryALTinterwordstretchfactor}{4}
\providecommand{\BIBentryALTinterwordspacing}{\spaceskip=\fontdimen2\font plus
\BIBentryALTinterwordstretchfactor\fontdimen3\font minus
  \fontdimen4\font\relax}
\providecommand{\BIBforeignlanguage}[2]{{%
\expandafter\ifx\csname l@#1\endcsname\relax
\typeout{** WARNING: IEEEtran.bst: No hyphenation pattern has been}%
\typeout{** loaded for the language `#1'. Using the pattern for}%
\typeout{** the default language instead.}%
\else
\language=\csname l@#1\endcsname
\fi
#2}}
\providecommand{\BIBdecl}{\relax}
\BIBdecl

\bibitem{Sepulveda2021}
N.~A. Sepulveda, J.~D. Jenkins, A.~Edington, S.~Mallapragada, Dharik, and R.~K.
  Lester, ``{The Design Space for Long-duration Energy Storage in Decarbonized
  Power Systems},'' \emph{Nature Energy}, p. Accepted, 2021.

\bibitem{Brown2020}
\BIBentryALTinterwordspacing
P.~R. Brown and A.~Botterud, ``{The Value of Inter-Regional Coordination and
  Transmission in Decarbonizing the US Electricity System},'' \emph{Joule}, 12
  2020. [Online]. Available:
  \url{https://linkinghub.elsevier.com/retrieve/pii/S2542435120305572}
\BIBentrySTDinterwordspacing

\bibitem{Guerra2020}
\BIBentryALTinterwordspacing
O.~J. Guerra, J.~Zhang, J.~Eichman, P.~Denholm, J.~Kurtz, and B.-M. Hodge,
  ``The value of seasonal energy storage technologies for the integration of
  wind and solar power,'' \emph{Energy {\&} Environmental Science}, vol.~13,
  no.~7, pp. 1909--1922, 2020. [Online]. Available:
  \url{https://doi.org/10.1039/d0ee00771d}
\BIBentrySTDinterwordspacing

\bibitem{MALLAPRAGADA2020115390}
\BIBentryALTinterwordspacing
D.~S. Mallapragada, N.~A. Sepulveda, and J.~D. Jenkins, ``Long-run system value
  of battery energy storage in future grids with increasing wind and solar
  generation,'' \emph{Applied Energy}, vol. 275, p. 115390, 2020. [Online].
  Available:
  \url{https://www.sciencedirect.com/science/article/pii/S0306261920309028}
\BIBentrySTDinterwordspacing

\bibitem{kuepper2020wind}
E.~Kuepper, H.~Teichgraeber, N.~Baumgaertner, A.~Bardow, and A.~Brandt, ``Wind
  data introduce error in time series reduction for capacity expansion
  modeling,'' \emph{Preparation}, vol. 123, p. 124, 2020.

\bibitem{teichgraeber2020extreme}
H.~Teichgraeber, C.~P. Lindenmeyer, N.~Baumg{\"a}rtner, L.~Kotzur, D.~Stolten,
  M.~Robinius, A.~Bardow, and A.~R. Brandt, ``Extreme events in time series
  aggregation: A case study for optimal residential energy supply systems,''
  \emph{Applied Energy}, vol. 275, p. 115223, 2020.

\bibitem{tejada2018enhanced}
D.~A. Tejada-Arango, M.~Domeshek, S.~Wogrin, and E.~Centeno, ``Enhanced
  representative days and system states modeling for energy storage investment
  analysis,'' \emph{IEEE Transactions on Power Systems}, vol.~33, no.~6, pp.
  6534--6544, 2018.

\bibitem{8017598}
Y.~Liu, R.~Sioshansi, and A.~J. Conejo, ``Hierarchical clustering to find
  representative operating periods for capacity-expansion modeling,''
  \emph{IEEE Transactions on Power Systems}, vol.~33, no.~3, pp. 3029--3039,
  2018.

\bibitem{9712383}
Y.~Yin, C.~He, T.~Liu, and L.~Wu, ``Risk-averse stochastic midterm schedule of
  thermal-hydro-wind system: A network-constrained clustered unit commitment
  approach,'' \emph{IEEE Transactions on Sustainable Energy}, pp. 1--1, 2022.

\bibitem{8536427}
C.~Feng, M.~Cui, B.-M. Hodge, S.~Lu, H.~F. Hamann, and J.~Zhang, ``Unsupervised
  clustering-based short-term solar forecasting,'' \emph{IEEE Transactions on
  Sustainable Energy}, vol.~10, no.~4, pp. 2174--2185, 2019.

\bibitem{8633397}
M.~A.~Z. Alvarez, K.~Agbossou, A.~Cardenas, S.~Kelouwani, and L.~Boulon,
  ``Demand response strategy applied to residential electric water heaters
  using dynamic programming and k-means clustering,'' \emph{IEEE Transactions
  on Sustainable Energy}, vol.~11, no.~1, pp. 524--533, 2020.

\bibitem{poncelet2016selecting}
K.~Poncelet, H.~H{\"o}schle, E.~Delarue, A.~Virag, and W.~D’haeseleer,
  ``Selecting representative days for capturing the implications of integrating
  intermittent renewables in generation expansion planning problems,''
  \emph{IEEE Transactions on Power Systems}, vol.~32, no.~3, pp. 1936--1948,
  2016.

\bibitem{KOLTSAKLIS2018563}
\BIBentryALTinterwordspacing
N.~E. Koltsaklis and A.~S. Dagoumas, ``State-of-the-art generation expansion
  planning: A review,'' \emph{Applied Energy}, vol. 230, pp. 563--589, 2018.
  [Online]. Available:
  \url{https://www.sciencedirect.com/science/article/pii/S0306261918312583}
\BIBentrySTDinterwordspacing

\bibitem{PONCELET2016631}
\BIBentryALTinterwordspacing
K.~Poncelet, E.~Delarue, D.~Six, J.~Duerinck, and W.~D’haeseleer, ``Impact of
  the level of temporal and operational detail in energy-system planning
  models,'' \emph{Applied Energy}, vol. 162, pp. 631--643, 2016. [Online].
  Available:
  \url{https://www.sciencedirect.com/science/article/pii/S0306261915013276}
\BIBentrySTDinterwordspacing

\bibitem{wogrin2014new}
S.~Wogrin, P.~Due{\~n}as, A.~Delgadillo, and J.~Reneses, ``A new approach to
  model load levels in electric power systems with high renewable
  penetration,'' \emph{IEEE Transactions on Power Systems}, vol.~29, no.~5, pp.
  2210--2218, 2014.

\bibitem{baumgartner2019rises3}
N.~Baumg{\"a}rtner, B.~Bahl, M.~Hennen, and A.~Bardow, ``Rises3: Rigorous
  synthesis of energy supply and storage systems via time-series relaxation and
  aggregation,'' \emph{Computers \& Chemical Engineering}, vol. 127, pp.
  127--139, 2019.

\bibitem{TEICHGRAEBER2022111984}
\BIBentryALTinterwordspacing
H.~Teichgraeber and A.~R. Brandt, ``Time-series aggregation for the
  optimization of energy systems: Goals, challenges, approaches, and
  opportunities,'' \emph{Renewable and Sustainable Energy Reviews}, vol. 157,
  p. 111984, 2022. [Online]. Available:
  \url{https://www.sciencedirect.com/science/article/pii/S1364032121012478}
\BIBentrySTDinterwordspacing

\bibitem{Lara2018}
\BIBentryALTinterwordspacing
C.~L. Lara, D.~S. Mallapragada, D.~J. Papageorgiou, A.~Venkatesh, and I.~E.
  Grossmann, ``{Deterministic electric power infrastructure planning:
  Mixed-integer programming model and nested decomposition algorithm},''
  \emph{European Journal of Operational Research}, vol. 271, no.~3, pp.
  1037--1054, 12 2018. [Online]. Available:
  \url{https://www.sciencedirect.com/science/article/pii/S0377221718304466}
\BIBentrySTDinterwordspacing

\bibitem{8316946}
K.~Kim, A.~Botterud, and F.~Qiu, ``Temporal decomposition for improved unit
  commitment in power system production cost modeling,'' \emph{IEEE
  Transactions on Power Systems}, vol.~33, no.~5, pp. 5276--5287, 2018.

\bibitem{Macqueen67somemethods}
J.~Macqueen, ``Some methods for classification and analysis of multivariate
  observations,'' in \emph{In 5-th Berkeley Symposium on Mathematical
  Statistics and Probability}, 1967, pp. 281--297.

\bibitem{scikit-learn}
F.~Pedregosa, G.~Varoquaux, A.~Gramfort, V.~Michel, B.~Thirion, O.~Grisel,
  M.~Blondel, P.~Prettenhofer, R.~Weiss, V.~Dubourg, J.~Vanderplas, A.~Passos,
  D.~Cournapeau, M.~Brucher, M.~Perrot, and E.~Duchesnay, ``Scikit-learn:
  Machine learning in {P}ython,'' \emph{Journal of Machine Learning Research},
  vol.~12, pp. 2825--2830, 2011.

\bibitem{10.5555/3001460.3001507}
M.~Ester, H.-P. Kriegel, J.~Sander, and X.~Xu, ``A density-based algorithm for
  discovering clusters in large spatial databases with noise,'' ser.
  KDD'96.\hskip 1em plus 0.5em minus 0.4em\relax AAAI Press, 1996, p.
  226–231.

\bibitem{Sheikholeslami1998WaveClusterAM}
G.~Sheikholeslami, S.~Chatterjee, and A.~Zhang, ``Wavecluster: A
  multi-resolution clustering approach for very large spatial databases,'' in
  \emph{VLDB}, 1998.

\bibitem{Fu2016PatternDF}
T.-C. Fu, F.~lai Chung, R.~W.~P. Luk, and Ng, ``Pattern discovery from stock
  time series using self-organizing maps,'' 2016.

\bibitem{Aghabozorgi2014}
\BIBentryALTinterwordspacing
S.~Aghabozorgi, T.~Y. Wah, T.~Herawan, H.~A. Jalab, M.~A. Shaygan, and
  A.~Jalali, ``A hybrid algorithm for clustering of time series data based on
  affinity search technique,'' \emph{The Scientific World Journal}, vol. 2014,
  pp. 1--12, 2014. [Online]. Available:
  \url{https://doi.org/10.1155/2014/562194}
\BIBentrySTDinterwordspacing

\bibitem{Kaufman1990}
\BIBentryALTinterwordspacing
L.~Kaufman and P.~J. Rousseeuw, Eds., \emph{Finding Groups in Data}.\hskip 1em
  plus 0.5em minus 0.4em\relax John Wiley {\&} Sons, Inc., Mar. 1990. [Online].
  Available: \url{https://doi.org/10.1002/9780470316801}
\BIBentrySTDinterwordspacing

\bibitem{10.5555/3000292.3000335}
E.~J. Keogh and M.~J. Pazzani, ``An enhanced representation of time series
  which allows fast and accurate classification, clustering and relevance
  feedback,'' in \emph{Proceedings of the Fourth International Conference on
  Knowledge Discovery and Data Mining}, ser. KDD'98.\hskip 1em plus 0.5em minus
  0.4em\relax AAAI Press, 1998, p. 239–243.

\bibitem{4761105}
V.~Hautamaki, P.~Nykanen, and P.~Franti, ``Time-series clustering by
  approximate prototypes,'' in \emph{2008 19th International Conference on
  Pattern Recognition}, 2008, pp. 1--4.

\bibitem{10.1145/331499.331504}
\BIBentryALTinterwordspacing
A.~K. Jain, M.~N. Murty, and P.~J. Flynn, ``Data clustering: A review,''
  \emph{ACM Comput. Surv.}, vol.~31, no.~3, p. 264–323, sep 1999. [Online].
  Available: \url{https://doi.org/10.1145/331499.331504}
\BIBentrySTDinterwordspacing

\bibitem{Keogh2003}
\BIBentryALTinterwordspacing
E.~Keogh and S.~Kasetty, ``On the need for time series data mining benchmarks:
  A survey and empirical demonstration,'' \emph{Data Mining and Knowledge
  Discovery}, vol.~7, no.~4, pp. 349--371, Oct 2003. [Online]. Available:
  \url{https://doi.org/10.1023/A:1024988512476}
\BIBentrySTDinterwordspacing

\bibitem{Bagnall2005}
\BIBentryALTinterwordspacing
A.~Bagnall and G.~Janacek, ``Clustering time series with clipped data,''
  \emph{Machine Learning}, vol.~58, no. 2-3, pp. 151--178, Feb. 2005. [Online].
  Available: \url{https://doi.org/10.1007/s10994-005-5825-6}
\BIBentrySTDinterwordspacing

\bibitem{doi:10.1137/1.9781611972726.12}
\BIBentryALTinterwordspacing
S.~Chu, E.~Keogh, D.~Hart, and M.~Pazzani, \emph{Iterative Deepening Dynamic
  Time Warping for Time Series}, pp. 195--212. [Online]. Available:
  \url{https://epubs.siam.org/doi/abs/10.1137/1.9781611972726.12}
\BIBentrySTDinterwordspacing

\bibitem{Wang2006}
\BIBentryALTinterwordspacing
X.~Wang, K.~Smith, and R.~Hyndman, ``Characteristic-based clustering for time
  series data,'' \emph{Data Mining and Knowledge Discovery}, vol.~13, no.~3,
  pp. 335--364, May 2006. [Online]. Available:
  \url{https://doi.org/10.1007/s10618-005-0039-x}
\BIBentrySTDinterwordspacing

\bibitem{Ratanamahatana2005}
\BIBentryALTinterwordspacing
C.~Ratanamahatana, E.~Keogh, A.~J. Bagnall, and S.~Lonardi, ``A novel bit level
  time series representation with implication of similarity search and
  clustering,'' in \emph{Advances in Knowledge Discovery and Data
  Mining}.\hskip 1em plus 0.5em minus 0.4em\relax Springer Berlin Heidelberg,
  2005, pp. 771--777. [Online]. Available:
  \url{https://doi.org/10.1007/11430919\_90}
\BIBentrySTDinterwordspacing

\bibitem{Lin2003}
\BIBentryALTinterwordspacing
J.~Lin, E.~Keogh, S.~Lonardi, and B.~Chiu, ``A symbolic representation of time
  series, with implications for streaming algorithms,'' in \emph{Proceedings of
  the 8th {ACM} {SIGMOD} workshop on Research issues in data mining and
  knowledge discovery - {DMKD} {\textquotesingle}03}.\hskip 1em plus 0.5em
  minus 0.4em\relax {ACM} Press, 2003. [Online]. Available:
  \url{https://doi.org/10.1145/882082.882086}
\BIBentrySTDinterwordspacing

\bibitem{Bagnall2006}
\BIBentryALTinterwordspacing
A.~Bagnall, C.~{\textquotedblleft}. Ratanamahatana, E.~Keogh, S.~Lonardi, and
  G.~Janacek, ``A bit level representation for time series data mining with
  shape based similarity,'' \emph{Data Mining and Knowledge Discovery},
  vol.~13, no.~1, pp. 11--40, May 2006. [Online]. Available:
  \url{https://doi.org/10.1007/s10618-005-0028-0}
\BIBentrySTDinterwordspacing

\bibitem{Shieh2008iSAXIA}
J.~Shieh and E.~J. Keogh, ``isax: indexing and mining terabyte sized time
  series,'' in \emph{KDD}, 2008.

\bibitem{TEICHGRAEBER20191283}
\BIBentryALTinterwordspacing
H.~Teichgraeber and A.~R. Brandt, ``Clustering methods to find representative
  periods for the optimization of energy systems: An initial framework and
  comparison,'' \emph{Applied Energy}, vol. 239, pp. 1283--1293, Apr. 2019.
  [Online]. Available: \url{https://doi.org/10.1016/j.apenergy.2019.02.012}
\BIBentrySTDinterwordspacing

\bibitem{TSO2020115190}
\BIBentryALTinterwordspacing
W.~W. Tso, C.~D. Demirhan, C.~F. Heuberger, J.~B. Powell, and E.~N.
  Pistikopoulos, ``A hierarchical clustering decomposition algorithm for
  optimizing renewable power systems with storage,'' \emph{Applied Energy},
  vol. 270, p. 115190, 2020. [Online]. Available:
  \url{https://www.sciencedirect.com/science/article/pii/S0306261920307029}
\BIBentrySTDinterwordspacing

\bibitem{7527691}
K.~Poncelet, H.~Höschle, E.~Delarue, A.~Virag, and W.~D’haeseleer,
  ``Selecting representative days for capturing the implications of integrating
  intermittent renewables in generation expansion planning problems,''
  \emph{IEEE Transactions on Power Systems}, vol.~32, no.~3, pp. 1936--1948,
  2017.

\bibitem{8442580}
A.~Almaimouni, A.~Ademola-Idowu, J.~Nathan~Kutz, A.~Negash, and D.~Kirschen,
  ``Selecting and evaluating representative days for generation expansion
  planning,'' in \emph{2018 Power Systems Computation Conference (PSCC)}, 2018,
  pp. 1--7.

\bibitem{8369128}
S.~Pineda and J.~M. Morales, ``Chronological time-period clustering for optimal
  capacity expansion planning with storage,'' \emph{IEEE Transactions on Power
  Systems}, vol.~33, no.~6, pp. 7162--7170, 2018.

\bibitem{hoffmann2020review}
M.~Hoffmann, L.~Kotzur, D.~Stolten, and M.~Robinius, ``A review on time series
  aggregation methods for energy system models,'' \emph{Energies}, vol.~13,
  no.~3, p. 641, 2020.

\bibitem{8334256}
D.~A. Tejada-Arango, M.~Domeshek, S.~Wogrin, and E.~Centeno, ``Enhanced
  representative days and system states modeling for energy storage investment
  analysis,'' \emph{IEEE Transactions on Power Systems}, vol.~33, no.~6, pp.
  6534--6544, 2018.

\bibitem{Fitiwi2015}
\BIBentryALTinterwordspacing
D.~Z. Fitiwi, F.~de~Cuadra, L.~Olmos, and M.~Rivier, ``{A new approach of
  clustering operational states for power network expansion planning problems
  dealing with RES (renewable energy source) generation operational variability
  and uncertainty},'' \emph{Energy}, vol.~90, no.~P2, pp. 1360--1376, 2015.
  [Online]. Available:
  \url{https://ideas.repec.org/a/eee/energy/v90y2015ip2p1360-1376.html}
\BIBentrySTDinterwordspacing

\bibitem{lythcke2016method}
C.~E. Lythcke-J{\o}rgensen, M.~M{\"u}nster, A.~V. Ensinas, and F.~Haglind, ``A
  method for aggregating external operating conditions in multi-generation
  system optimization models,'' \emph{Applied Energy}, vol. 166, pp. 59--75,
  2016.

\bibitem{sun2019data}
M.~Sun, F.~Teng, X.~Zhang, G.~Strbac, and D.~Pudjianto, ``Data-driven
  representative day selection for investment decisions: A cost-oriented
  approach,'' \emph{IEEE Transactions on Power Systems}, vol.~34, no.~4, pp.
  2925--2936, 2019.

\bibitem{LI2022107697}
\BIBentryALTinterwordspacing
C.~Li, A.~J. Conejo, J.~D. Siirola, and I.~E. Grossmann, ``On representative
  day selection for capacity expansion planning of power systems under extreme
  operating conditions,'' \emph{International Journal of Electrical Power \&
  Energy Systems}, vol. 137, p. 107697, 2022. [Online]. Available:
  \url{https://www.sciencedirect.com/science/article/pii/S0142061521009248}
\BIBentrySTDinterwordspacing

\bibitem{8610317}
M.~Sun, F.~Teng, X.~Zhang, G.~Strbac, and D.~Pudjianto, ``Data-driven
  representative day selection for investment decisions: A cost-oriented
  approach,'' \emph{IEEE Transactions on Power Systems}, vol.~34, no.~4, pp.
  2925--2936, 2019.

\bibitem{Tavakoli2020}
\BIBentryALTinterwordspacing
N.~Tavakoli, S.~Siami-Namini, M.~A. Khanghah, F.~M. Soltani, and A.~S. Namin,
  ``An autoencoder-based deep learning approach for clustering time series
  data,'' \emph{{SN} Applied Sciences}, vol.~2, no.~5, Apr. 2020. [Online].
  Available: \url{https://doi.org/10.1007/s42452-020-2584-8}
\BIBentrySTDinterwordspacing

\bibitem{Guo2017}
\BIBentryALTinterwordspacing
X.~Guo, X.~Liu, E.~Zhu, and J.~Yin, ``Deep clustering with convolutional
  autoencoders,'' in \emph{Neural Information Processing}.\hskip 1em plus 0.5em
  minus 0.4em\relax Springer International Publishing, 2017, pp. 373--382.
  [Online]. Available: \url{https://doi.org/10.1007/978-3-319-70096-0\_39}
\BIBentrySTDinterwordspacing

\bibitem{8663347}
M.~Khodayar, S.~Mohammadi, M.~E. Khodayar, J.~Wang, and G.~Liu, ``Convolutional
  graph autoencoder: A generative deep neural network for probabilistic
  spatio-temporal solar irradiance forecasting,'' \emph{IEEE Transactions on
  Sustainable Energy}, vol.~11, no.~2, pp. 571--583, 2020.

\bibitem{taylor2015convex}
\BIBentryALTinterwordspacing
J.~Taylor, \emph{Convex Optimization of Power Systems}, ser. Convex
  Optimization of Power Systems.\hskip 1em plus 0.5em minus 0.4em\relax
  Cambridge University Press, 2015. [Online]. Available:
  \url{https://books.google.com/books?id=JBdoBgAAQBAJ}
\BIBentrySTDinterwordspacing

\bibitem{Jenkins2017}
\BIBentryALTinterwordspacing
J.~Jenkins and N.~Sepulveda, ``{Enhanced Decision Support for a Changing
  Electricity Landscape: the GenX Configurable Electricity Resource Capacity
  Expansion Model},'' MIT Energy Initiative, Tech. Rep., 2017. [Online].
  Available:
  \url{https://energy.mit.edu/wp-content/uploads/2017/10/Enhanced-Decision-Support-for-a-Changing-Electricity-Landscape.pdf}
\BIBentrySTDinterwordspacing

\bibitem{Knueven2020}
\BIBentryALTinterwordspacing
B.~Knueven, J.~Ostrowski, and J.-P. Watson, ``On mixed-integer programming
  formulations for the unit commitment problem,'' \emph{{INFORMS} Journal on
  Computing}, Jun. 2020. [Online]. Available:
  \url{https://doi.org/10.1287/ijoc.2019.0944}
\BIBentrySTDinterwordspacing

\bibitem{https://doi.org/10.48550/arxiv.2203.12426}
\BIBentryALTinterwordspacing
M.~Barbar, D.~S. Mallapragada, and R.~Stoner, ``Impact of demand growth on
  decarbonizing india's electricity sector and the role for energy storage,''
  2022. [Online]. Available: \url{https://arxiv.org/abs/2203.12426}
\BIBentrySTDinterwordspacing

\bibitem{Goodfellow-et-al-2016}
I.~Goodfellow, Y.~Bengio, and A.~Courville, \emph{Deep Learning}.\hskip 1em
  plus 0.5em minus 0.4em\relax MIT Press, 2016,
  \url{http://www.deeplearningbook.org}.

\bibitem{chollet2015keras}
F.~Chollet \emph{et~al.}, ``Keras,'' \url{https://github.com/fchollet/keras},
  2015.

\bibitem{JMLR:v15:srivastava14a}
\BIBentryALTinterwordspacing
N.~Srivastava, G.~Hinton, A.~Krizhevsky, I.~Sutskever, and R.~Salakhutdinov,
  ``Dropout: A simple way to prevent neural networks from overfitting,''
  \emph{Journal of Machine Learning Research}, vol.~15, no.~56, pp. 1929--1958,
  2014. [Online]. Available:
  \url{http://jmlr.org/papers/v15/srivastava14a.html}
\BIBentrySTDinterwordspacing

\bibitem{pmlr-v37-ioffe15}
\BIBentryALTinterwordspacing
S.~Ioffe and C.~Szegedy, ``Batch normalization: Accelerating deep network
  training by reducing internal covariate shift,'' in \emph{Proceedings of the
  32nd International Conference on Machine Learning}, ser. Proceedings of
  Machine Learning Research, F.~Bach and D.~Blei, Eds., vol.~37.\hskip 1em plus
  0.5em minus 0.4em\relax Lille, France: PMLR, 07--09 Jul 2015, pp. 448--456.
  [Online]. Available: \url{https://proceedings.mlr.press/v37/ioffe15.html}
\BIBentrySTDinterwordspacing

\bibitem{MacKay2003}
D.~MacKay, \emph{Information Theory, Inference and Learning Algorithms}.\hskip
  1em plus 0.5em minus 0.4em\relax Cambridge University Press, 2003, ch.
  Chapter 20. An Example Inference Task: Clustering.

\bibitem{MALLAPRAGADA20181231}
\BIBentryALTinterwordspacing
D.~S. Mallapragada, D.~J. Papageorgiou, A.~Venkatesh, C.~L. Lara, and I.~E.
  Grossmann, ``Impact of model resolution on scenario outcomes for electricity
  sector system expansion,'' \emph{Energy}, vol. 163, pp. 1231--1244, 2018.
  [Online]. Available:
  \url{https://www.sciencedirect.com/science/article/pii/S0360544218315238}
\BIBentrySTDinterwordspacing

\bibitem{ercot_load}
{Electric Reliability Council of Texas}, ``Hourly load data archive,''
  \url{https://www.ercot.com/gridinfo/load/load\_hist}.

\bibitem{ercot_weather}
{Rojowsky, K., Gothandaraman, A. \& Beaucage, P.}, ``Hourly wind and solar
  generation profiles (1980-2019),'' Tech. Rep., 2020.

\bibitem{ercot_network}
{Pacific Northwest National Laboratory}, ``{8-Bus ERCOT Bulk System Models},''
  \url{https://github.com/pnnl/tesp}.

\bibitem{ATB2020}
{National Renewable Energy Laboratory (NREL)}, ``{Annual Technology Baseline:
  Electricity},'' 2020.

\bibitem{reuther2018interactive}
A.~Reuther, J.~Kepner, C.~Byun, S.~Samsi, W.~Arcand, D.~Bestor, B.~Bergeron,
  V.~Gadepally, M.~Houle, M.~Hubbell, M.~Jones, A.~Klein, L.~Milechin,
  J.~Mullen, A.~Prout, A.~Rosa, C.~Yee, and P.~Michaleas, ``Interactive
  supercomputing on 40,000 cores for machine learning and data analysis,'' in
  \emph{2018 IEEE High Performance extreme Computing Conference (HPEC)}.\hskip
  1em plus 0.5em minus 0.4em\relax IEEE, 2018, pp. 1--6.

\bibitem{https://doi.org/10.1002/sam.11161}
\BIBentryALTinterwordspacing
A.~Zimek, E.~Schubert, and H.-P. Kriegel, ``A survey on unsupervised outlier
  detection in high-dimensional numerical data,'' \emph{Statistical Analysis
  and Data Mining: The ASA Data Science Journal}, vol.~5, no.~5, pp. 363--387,
  2012. [Online]. Available:
  \url{https://onlinelibrary.wiley.com/doi/abs/10.1002/sam.11161}
\BIBentrySTDinterwordspacing

\bibitem{optmodel}
\BIBentryALTinterwordspacing
M.~Barbar, ``Source code for: Optimization model,'' \emph{Github}.
  \url{https://github.mit.edu/mbarbar/phd-thesis/tree/master/optmodel}, 2022.
  [Online]. Available:
  \url{\url{https://github.mit.edu/mbarbar/phd-thesis/tree/master/optmodel}}
\BIBentrySTDinterwordspacing

\bibitem{autoencoder}
\BIBentryALTinterwordspacing
------, ``Source code for: Time-series clustering autoencoder,'' \emph{Github}.
  \url{https://github.mit.edu/mbarbar/phd-thesis/tree/master/autoencoder},
  2022. [Online]. Available:
  \url{\url{https://github.mit.edu/mbarbar/phd-thesis/tree/master/autoencoder}}
\BIBentrySTDinterwordspacing

\bibitem{marc_barbar_2022_6484347}
\BIBentryALTinterwordspacing
M.~Barbar and D.~S. Mallapragada, ``{Representative period selection for power
  system planning using autoencoder-based dimensionality reduction},'' Apr.
  2022. [Online]. Available: \url{https://doi.org/10.5281/zenodo.6484347}
\BIBentrySTDinterwordspacing

\end{thebibliography}

\end{document}